\documentclass[11pt]{article}

\usepackage[final]{acl}

\usepackage{times}
\usepackage{latexsym}
\usepackage{booktabs}
\usepackage{pifont}
\usepackage{xcolor,colortbl}
\usepackage{threeparttable}
\usepackage[table]{xcolor}
\usepackage{multirow} 
\usepackage{amsmath}
\usepackage{booktabs} 
\usepackage{soul}

\sethlcolor{blue!12}

\usepackage[T1]{fontenc}

\usepackage[utf8]{inputenc}

\usepackage{microtype}

\usepackage{inconsolata}

\usepackage{graphicx}

\usepackage{xspace}

\title{From Proof to Program: Characterizing Tool-Induced Reasoning Hallucinations in Large Language Models}

\author{
 \textbf{Farima Fatahi Bayat},
 \textbf{Pouya Pezeshkpour},
 \textbf{Estevam Hruschka}
\\
 Megagon Labs
\\
    \{farima, pouya, estevam\}@megagon.ai
\\
}

\def \dataset{\textsc{PyMath}\xspace}
\def \task{\textsc{TIM}\xspace}

\newcommand{\postspace}{\vskip -3mm}
\newcommand{\minipostspace}{\vskip -1mm}

\definecolor{cadmiumgreen}{rgb}{0.0, 0.42, 0.24}
\definecolor{cardinal}{rgb}{0.77, 0.12, 0.23}
\definecolor{cadmiumred}{rgb}{0.89, 0.0, 0.13}

\usepackage[most]{tcolorbox}
\newtcolorbox[list inside=prompt,auto counter,number within=section]{prompt}[1][]{
    fontupper=\ttfamily\footnotesize,
    boxsep=5pt,
    left=0pt,
    right=0pt,
    top=0pt,
    bottom=0pt,
    boxrule=1pt,
    breakable,
    #1,
}

\newtcolorbox{example}[2][]{%
  enhanced, breakable,
  colback=white,           
  colframe=black!10,
  boxrule=0.4pt,
  coltitle=black,
  fonttitle=\bfseries,
  title={#2},
  #1
}

\begin{document}
\maketitle
\begin{abstract}

Tool-augmented Language Models (TaLMs) can invoke external tools 
to solve 
problems beyond their parametric capacity. However, it remains unclear whether these tool-enabled gains reflect trustworthy reasoning.
Focusing on the Code Interpreter tool,
we show that even when tools are selected and executed correctly, TaLMs treat tool outputs as substitutes for reasoning, producing solutions that appear correct but lack coherent justification. We term this failure mode \textbf{T}ool-\textbf{I}nduced \textbf{M}yopia (\textbf{\task{}}), and study it using
\dataset, 
a benchmark of 1,679 competition-level mathematical problems for which
Python code is \textit{helpful but not sufficient}. We further develop a multi-dimensional evaluation suite to quantify reasoning degradation in TaLMs relative to their non-tool counterparts. Our findings reveal that while TaLMs achieve up to a 19.3 \textit{percentage point} gain in final-answer accuracy, their reasoning behavior consistently deteriorates (e.g., non-tool language models win up to 41.5\% more often in pairwise comparisons of reasoning processes). 
This degradation intensifies with tool use; the more frequently a model invokes tools, the less coherent its reasoning becomes. Moreover, tool use shifts errors from arithmetic mistakes toward global reasoning failures (logic, assumption, creativity).
Finally, we propose a preference-optimization-based framework that realigns TaLMs to use tool outputs as assistive evidence, improving both final-answer accuracy and the reasoning depth under tool use.\footnote{Code and data are available at: \url{https://github.com/megagonlabs/TIM}.}

\end{abstract}

\section{Introduction}

Large Language Models (LLMs) have grown increasingly capable, yet relying solely on their parametric knowledge introduces key limitations, including the inability to access real-time or domain-specific information~\citep{yu-ji-2024-information, wang2025actingreasoningmoreteaching}, perform precise computations~\citep{lu-etal-2023-survey}, or fully comprehend user intentions~\citep{qian-etal-2024-tell}. 
To address these shortcomings, Tool-augmented Reasoning (TIR)~\citep{toolformer, gou2024tora} has emerged as a promising paradigm. It enables LLMs to integrate natural language reasoning with external tools. 
Frontier LLMs \citep{gptmodels, geminimodels, claudemodels} now offer native, sandboxed execution for select tools, choosing when to call, executing, and integrating results. 


While tool calling significantly extends LLMs' utility in computation, retrieval, and procedural tasks, it also introduces new sources of failures. The most fundamental errors arise from \textit{tool hallucinations}, where models either select inappropriate tools or misuse them, leading to incorrect or irrelevant outputs~\citep{10.5555/3737916.3741936, xu2025reducingtoolhallucinationreliability}. 
However, even with correct tool selection and successful execution, LLMs can still generate non-factual outputs or flawed reasoning. Prior work ties factual errors to conflicts between parametric and retrieved knowledge~\citep{sun2025redeepdetectinghallucinationretrievalaugmented} as well as to the propagation of errors from retrieved content~\citep{magesh2024hallucinationfreeassessingreliabilityleading}. Additional studies show that LLMs' unrestricted access to external tools can induce tool overuse ~\citep{qian-etal-2025-smart}, which confuses the model and harms performance. It also encourages cognitive offloading \citep{wang2025actingreasoningmoreteaching}, which limits the model’s use of its internal reasoning capabilities. 
Despite these observations, it remains unclear whether tool-augmented reasoning is hallucination-free even when other failure modes are controlled.

\begin{figure*}
    \centering
    \includegraphics[width=0.98\linewidth]{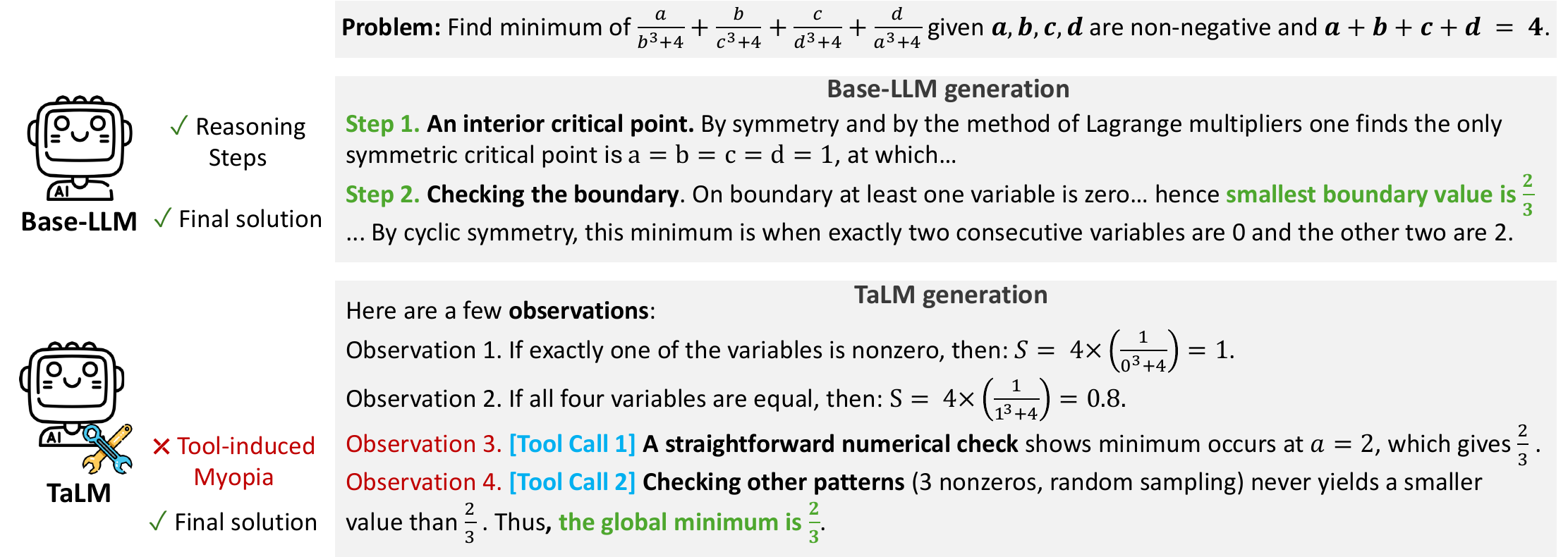}
    \caption{Comparison of Base LLM and Tool-augmented LLM (TaLM) reasoning. The Base LLM (top) derives the solution through step-by-step mathematical reasoning, while the TaLM (bottom) relies on \textbf{empirical checks} and multiple tool calls to search for the minimum, a failure mode characteristic of Tool-Induced Myopia (\task). 
    }
    \label{fig:example}
    \postspace
    \minipostspace
\end{figure*}

In this work, we focus exclusively on the Code Interpreter tool for mathematical problem solving to ensure strictly controlled tool use, with correct invocation and error-free execution. Our design avoids confounds introduced by multi-tool settings, such as API failures, interface mismatches, retrieval drift, or external data quality issues~\citep{zhong2025complexfuncbench, faghih2025tool, maekawa2025towards, han2025search}, thereby enabling a controlled investigation. Despite these idealized conditions, we identify a new class of hallucinations, which we term \textbf{T}ool-\textbf{I}nduced \textbf{M}yopia (\task): a failure mode where access to an external tool (e.g., a Code Interpreter) causes the model to narrow its reasoning to what the tool can compute, rather than leveraging its full internal reasoning capabilities. \task{} erodes user trust, as correct final answers can conceal flawed, tool-dependent reasoning.


Figure~\ref{fig:example} illustrates this behavior: with Code Interpreter access (TaLM response), the model repeatedly performs \textbf{empirical checks} instead of producing the required reasoning steps generated by the same model without tools (Base-LLM response). Although the tool is correctly invoked and returns valid outputs, the model’s reasoning depth is reduced. Importantly, existing evaluation approaches fail to capture this failure mode. Final-answer accuracy cannot detect it, since both responses are correct, and step-level logical-consistency metrics \citep{lightman2023letsverifystepstep, zheng-etal-2025-processbench, xiareasoneval, liu2025enhancingmathematicalreasoninglarge} also fail because the reasoning remains superficially coherent despite skipping essential logical steps. As a result, exposing \task requires richer, multi-dimensional evaluation of how models reason under tool use.

To investigate \task, we introduce \dataset, a dataset comprising 1,679 competition-level mathematical problems collected from multiple sources and curated to elicit, measure, and mitigate \task under Code Interpreter access. We specifically target problems for which code-based computation is helpful but not sufficient for a complete solution
(details in Section~\ref{sec:data_curation}). We then quantify reasoning degradation in TaLMs compared to their Base (non-tool) counterparts using a comprehensive suite of reference-free and reference-based metrics (Section~\ref{sec:evaluation_suite}). Our results show that, even when TaLMs produce correct final answers (up to a 19.3 \textit{percentage point} in final performance gains), they often overrely on tool outputs and produce shallower reasoning traces than non-tool models.
This degradation intensifies with tool use: as models invoke tools more frequently, their reasoning becomes less coherent. The associated errors also shift from arithmetic mistakes to global reasoning failures involving logic, assumption, and creativity. Finally, our manual audit finds evidence of \task{} in \textasciitilde{}55\% of high-risk model-generated solutions.

To mitigate this issue, we propose two complementary strategies: (1) a prompting intervention that encourages models to treat tools as reasoning aids, and (2) a DPO-based \citep{10.5555/3666122.3668460} preference-optimization that aligns TaLMs to integrate the Code Interpreter as an assistant that supports, rather than replaces, mathematical reasoning. 
On the evaluation split of \dataset, the fine-tuned TaLM demonstrates improved reasoning behavior and surpasses both the vanilla TaLM (+0.6\%) and the Base LLM (+3.0\%) in final-answer accuracy.
In summary, our key contributions are:
\begin{itemize}
\item \textbf{Tool-Induced Myopia}: We identify \task, a new class of TaLM hallucinations where tool access reduces reasoning depth even under idealized tool-use conditions.
\item \textbf{Benchmark and Evaluation}: We introduce \dataset{}, a competition-level math benchmark with open-ended solutions, along with a multi-dimensional evaluation suite that reveals when and why \task{} emerges and guides safer tool integration.
\item \textbf{Mitigation}: We propose a training-free prompting method and a preference-optimization fine-tuning approach that reduce \task and improve reasoning depth in TaLMs.
\end{itemize}

\section{Tool-induced Myopia}
We define \textbf{Tool-Induced Myopia (\task{})} as a type of hallucination that arises in a controlled tool-use setting where the task requires substantial reasoning beyond what the tool alone can provide. In particular, we study competition-style mathematical problem solving with access to a Code Interpreter under two assumptions: (i) tool invocation is correct, and (ii) execution is error-free.
In practice, it substitutes enumeration for proof, skips necessary derivations, mistakes empirical checks for universal guarantees (e.g., brute-force search), and may prematurely stop once code returns a plausible output. 
Note that using code solely for precise computation (e.g., evaluating a determinant or numerically finding a root), while also providing the necessary derivations, is \textbf{not} considered \task{}. 
Figure \ref{fig:example} illustrates \task{}: the Base-LLM solves the problem through step-by-step mathematical reasoning, whereas the TaLM shifts to an exploration-based approach, using empirical observations from internal computation and Code Interpreter to reach the correct answer through numerical search. \task{} degrades reasoning by overrelying on tool outputs in place of the reasoning needed to justify the full solution.
Notably, the TaLM in Figure~\ref{fig:example} still produces the correct final answer via a logically coherent yet incomplete sequence of steps. 
To expose \task{}, we next introduce an evaluation benchmark and a suite of targeted metrics.

\section{Evaluating \task in Language Models}
\subsection{\dataset{}}
\label{sec:data_curation}
We focus on the domain of mathematical problem solving, where reasoning and computation are tightly coupled, and evaluate LLMs under two settings: with and without access to a Python Code Interpreter. We collect English, text-only competition-level math problems from multiple sources, summarized in Table~\ref{tab:data_sources_and_stats}. To mitigate data contamination, we restrict AIME \citep{AIME20242025} to 2024–2025 problems. From Omni-Math \citep{gao2024omnimathuniversalolympiadlevel}, which provides difficulty ratings (1-10 scale), we retain only problems with difficulty $\geq5$, as frontier LLMs have already saturated performance on easier problems \citep{gpt5, claudeopus4andsonnet4}. Finally, we use the full set of problems from OlympiadBench \citep{he-etal-2024-olympiadbench} and OlympicArena \citep{huang2024olympicarena}.

To elicit \task{} in TaLMs more effectively, we target problems for which Python code is \textit{helpful but not sufficient}. This setting creates a natural opportunity for tool use, while still \emph{requiring} reasoning over tool outputs to derive a solution. To identify such cases, we adopt an LLM-as-a-judge protocol~\citep{gu2025surveyllmasajudge} to assess each problem along two dimensions: \textbf{(1) Python Usefulness}: whether Python code helps solve the problem; and \textbf{(2) Python Sufficiency}: whether Python code alone, without additional LLM reasoning, is sufficient to fully solve the problem. Appendix~\ref{appendix:data_curation} provides the filtering prompt and a manual validation of the judge.

Table~\ref{tab:data_sources_and_stats} summarizes the resulting statistics by source. Our final dataset comprises 1,679 competition-level problems with step-by-step reference solutions: 1,000 problems (\textasciitilde{}60\%, distributed across sources and problem difficulties) for evaluation, and the rest for TaLM training (10\% development, 90\% DPO fine-tuning; see Section~\ref{sec:talm_tuning}).



\begin{table}[t]
\centering
\small
\begin{tabular}{lccc}
\toprule
\textbf{Source} & \textbf{Total} & \textbf{U but not S} & \textbf{Eval (\%)} \\
\midrule
AIME$^{*}$& 60 & 13 &  23.1\\
OlympiadBench& 674 & 171 &  52.0\\
OlympicArena& 169 & 47 &  12.8\\
Omni-Math& 2569 & 1448 &  62.3 \\
\rowcolor{green!10}
\textbf{\dataset} & - & \textbf{1679} &  \textbf{59.5}\\
\bottomrule
\end{tabular}
\caption{Sources and statistics of competition-level math problems included in the \dataset{} benchmark (containing problems where Python is Useful but not Sufficient). $^*$Only AIME 2024–2025 problems are included.}
\label{tab:data_sources_and_stats}
\postspace
\end{table}


\subsection{Evaluation Suite}
\label{sec:evaluation_suite}
Evaluating LLMs' mathematical reasoning has traditionally relied on outcome-based evaluation, most predominantly final-answer accuracy~\citep{10.1145/3773985, ahn-etal-2024-large}. However, recent studies~\citep{mondorf2024accuracyevaluatingreasoningbehavior, yee2024faithful} have shown that LLMs can reach a correct final answer despite flawed reasoning.
This gap has motivated a shift toward process-based evaluation, which assesses LLMs' \textit{reasoning behavior} rather than their task performance.

Process-based approaches span two major families. \textit{Reference-free} methods do not rely on gold reasoning traces and instead assess reasoning behavior via mechanisms such as self-consistency \cite{liu2025enhancingmathematicalreasoninglarge}, or pairwise comparison (e.g., win rate \citep{chen2025learningreasonfactuality, kim_2024_toolllm}). 
In contrast, \textit{reference-dependent} approaches evaluate reasoning against a gold step-by-step solution, enabling fine-grained detection of invalid, missing, or inconsistent steps \citep{yan2025verifybench, chernyshev2025u}.
A recent line of work extends this idea via process reward models (PRMs) to score partial solutions step by step, providing a learned measure of reasoning soundness without requiring gold supervision~\citep{lightman2023letsverifystepstep, zheng-etal-2025-processbench, li-etal-2025-fg}. 

As illustrated in Figure~\ref{fig:example}, neither final-answer accuracy nor unidimensional process-based metrics (e.g., PRM scores alone) are sufficient to expose \task in TaLMs: the model may output the correct answer and generate seemingly coherent reasoning steps while still exhibiting degraded reasoning under tool use. This motivates the need for a multi-dimensional evaluation suite that jointly captures: \textbf{(i)} task outcome, \textbf{(ii)} counterfactual impact of tool use, \textbf{(iii)} divergence from ground truth reasoning traces, and \textbf{(iv)} step-level logical consistency. Next, we introduce these four evaluation dimensions and clarify why each one is necessary, yet insufficient on its own, for diagnosing \task.

\paragraph{Final-Answer Accuracy:}
\label{sec:final_ans_acc}
We first measure final-answer accuracy as the standard measure of task success.  
Although this metric cannot reveal whether a TaLM solved the problem through mathematical reasoning or tool-driven shortcuts, it confirms that \task failure affects reasoning, not task performance. 
For all subsequent metrics which target reasoning behavior, we evaluate models only on problems with correct final answer to isolate the effect of tool use on how the model reasons, rather than whether it succeeds.

\paragraph{Win Rate:}
\label{sec:win_rate}
To assess whether tool use meaningfully affects reasoning behavior, we compare solutions from a TaLM to those of the corresponding Base (non-tool) LLM. Following prior work showing that LLMs can reliably approximate human preferences~\citep{10.5555/3666122.3668142, chen2025learningreasonfactuality}, we use an LLM judge with a rubric adapted from~\citet{petrov2025proofbluffevaluatingllms} covering four error types: (1) \textit{logic errors} (logical fallacies or unjustified leaps), (2) \textit{assumption errors} (unsupported or incorrect assumptions), (3) \textit{creativity errors} (invalid solution strategies), and (4) \textit{algebra/arithmetic errors} (critical symbolic or numeric mistakes). 
To avoid order bias, we randomize the presentation order of TaLM and Base-LLM solutions~\citep{shi2025judgingjudgessystematicstudy}.
The judge then selects the solution with fewer errors as the winner (prompt in Appendix~\ref{appendix:winrate_prompt}).


\textit{Win Rate} measures the proportion of comparisons in which the TaLM solution is preferred (and vice versa). A lower Win Rate indicates tool access \emph{harms} the LLM's reasoning behavior. This comparison-based setup reduces single-judge bias~\citep{liu-etal-2025-verifact, fatahi-bayat-etal-2025-factbench} and captures the counterfactual effect of tool use.

\paragraph{Miss Rate:}
\label{sec:miss_rate}
Inspired by recall-based measures in long-form factuality evaluation \citep{safe, liu-etal-2025-verifact}, we define \textit{Miss Rate} as the proportion of reasoning steps in the ground-truth solution that are absent from the model-generated solution, i.e., $\frac{|\text{missing steps}|}{|\text{gold steps}|}$. A high Miss Rate indicates that the model abandoned a valid derivation, skipped necessary steps, or replaced reasoning with trial-and-error code execution. Unlike final-answer accuracy, Miss Rate detects invalid reasoning paths even when the model reaches the correct solution. However, this metric may over-penalize a solution when the reference solution(s) do not cover all valid reasoning paths (prompt in Appendix~\ref{appendix:missrate_prompt}).


\paragraph{PRM Accuracy:}
\label{sec:prm}
Finally, we evaluate step-level reasoning behavior using a Process Reward Model, specifically \textsc{Qwen2.5-Math-7B-PRM800K}~\citep{zheng-etal-2025-processbench}, trained on $\sim$800K annotated mathematical steps. We aggregate step-level scores to obtain an overall correctness score for each solution. However, while PRMs evaluate step correctness, they may fail to detect reasoning shortcuts where code outputs replace derivations, as they lack a holistic view of the solution. In our analysis, we use PRM scores primarily to compare Base-LLM and TaLM solutions, and Appendix~\ref{appendix:prm_stability} shows that this comparative signal remains stable under moderate perturbations. 

These four metrics jointly provide minimal yet complete coverage of \task{}: Final-answer accuracy captures outcomes; Win Rate isolates the tool's counterfactual impact on reasoning behavior; Miss Rate quantifies divergence from valid reference reasoning; and a PRM assesses step-level soundness without relying on the ground truth. This set is sufficient because \task{} exhibits a clear and convergent pattern: final-answer accuracy stays the same or improves in TaLMs, while Win Rate declines, Miss Rate increases, and PRM accuracy drops. This suite both distinguishes \task{} from alternative explanations (e.g., general model weakness or genuinely helpful tool use) and is falsifiable: when tools are removed, the pattern recedes, and when tool reliance is increased, it strengthens. Together, these properties can establish both the existence and mechanism of Tool-Induced Myopia.

\section{Experimental Setup}
\label{sec:experimental_setup}
\textbf{Large Language Models:} We evaluate frontier proprietary language models equipped with in-house Code Interpreters. These models can autonomously decide when to invoke tools, allowing us to study tool-augmented reasoning without modifying or intervening in their internal execution pipeline.
We benchmark seven models across three families:
(1) OpenAI GPT-4.1-mini and GPT-4.1 (non-thinking LLMs; \citealp{gpt4.1}), and o4-mini~\citep{o4-mini-model} and GPT-5~\citep{gpt5} (thinking LLMs), (2) Gemini-2.0-Flash (non-thinking; \citealp{gemini-2-model}) and Gemini-2.5-Flash (thinking; \citealp{gemini2.5-flash}), (3) Claude-Opus-4 (thinking; \citealp{claudeopus4andsonnet4}).
%

\noindent
\textbf{Evaluation Benchmark}: We use the evaluation split of \dataset as our benchmark dataset and apply our four-dimensional evaluation suite to measure reasoning and tool-use behavior. 
We use GPT-5 as the judge for all LLM-as-a-judge evaluations.

\begin{table*}[t]
\centering
\small
\renewcommand{\arraystretch}{0.9}
\begin{tabular}{l l c c c c}
\toprule
\textbf{Model} & \textbf{Variant} & \textbf{Final Acc. ($\uparrow$)} & \textbf{Miss Rate ($\downarrow$)} & \textbf{Win Rate ($\uparrow$)} & \textbf{PRM Acc. ($\uparrow$)} \\
\midrule
\multirow{2}{*}{GPT-4.1-mini} 
 & Base & 30.0 & 45.7 & \colorbox{green!10}{58.6} & \colorbox{green!10}{93.0} \\
 & TaLM & 28.7 & 49.9 & 41.4 & 88.9 \\
\midrule
\multirow{2}{*}{GPT-4.1} 
 & Base & 24.6 & 48.1 & 54.4 & 88.6 \\
 & TaLM & 27.0 & 49.9 & 45.6 & 85.9 \\
\midrule
\multirow{2}{*}{o4-mini} 
 & Base & 45.1 & 45.5 & 49.0 & 73.6 \\
 & TaLM & 64.4 & 47.6 & 50.9 & 67.9 \\
\midrule
\multirow{2}{*}{GPT-5-Thinking} 
 & Base & 67.5 & \colorbox{green!10}{38.8} & 56.0 & 57.5 \\
 & TaLM & \colorbox{green!10}{71.9} & 43.8 & 44.0 & 50.2 \\
\midrule
\multirow{2}{*}{Gemini-2.0-Flash} 
 & Base & 24.3 & 54.2 & 52.7 & 65.0 \\
 & TaLM & 25.1 & 56.6 & 47.3 & 68.5 \\
\midrule
\multirow{2}{*}{Gemini-2.5-Flash} 
 & Base & 45.4 & 40.2 & 54.6 & 81.5 \\
 & TaLM & 45.7 & 40.9 & 45.4 & 78.8 \\
\midrule
\multirow{2}{*}{Claude-Opus-4} 
 & Base & 28.0 & 50.9 & 41.3 & 77.9 \\
 & TaLM & 40.5 & 52.8 & \colorbox{green!10}{58.6} & 57.8 \\
 \midrule
\multirow{2}{*}{\textbf{Average}} & Base & 37.8 & \textbf{46.2} & \textbf{52.4} & \textbf{76.7} \\
                                 & TaLM & \textbf{43.3} & 48.8 & 47.6 & 71.1 \\
\bottomrule
\end{tabular}
\caption{Performance of Base-LLMs and TaLMs across four evaluation metrics. Best score in each column is highlighted in \colorbox{green!10}{green}. On average, TaLMs achieve higher Final-answer Accuracy (Final Acc.), yet Base-LLMs exhibit greater reasoning depth, confirming the presence of \task{}.
}
\vspace{2mm}
\label{tab:main_results}
\postspace
\end{table*}

\section{Results and Analyses}

In this section, we first evaluate Base-LLMs and their tool-augmented counterparts (TaLMs) using our four-dimensional evaluation suite (Section~\ref{sec:benchmarking_results}). We then examine how the severity of \task{} varies with the number of tool calls (Section~\ref{sec:tool_call_and_tim}) and analyze whether the complexity of TaLM-generated code correlates with reasoning hallucinations (Appendix ~\ref{appendix:code_complexity}).
Next, we apply the error taxonomy from Section~\ref{sec:evaluation_suite} to characterize error types introduced by tool access (Section~\ref{sec:error_type_analysis}), and analyze tool invocation frequency given that \dataset{} encourages, but does not guarantee, tool use (Section~\ref{sec:tool_invocation_rate}).
Finally, Section~\ref{sec:quality_analysis} presents a qualitative analysis confirming that solutions flagged as high risk by our metrics indeed exhibit \task{} behavior.

\subsection{Base-LLMs Show Stronger Reasoning Despite Lower Accuracy}
\label{sec:benchmarking_results}
Our benchmarking results on the evaluation set of \dataset are presented in Table~\ref{tab:main_results}. We report Final-answer Accuracy, Miss Rate, Win Rate, and PRM Accuracy across the Base and TaLM variants of seven LLMs. The results reveal a consistent pattern characteristic of \task{}: although \textbf{TaLMs achieve higher Final-answer Accuracy, Base-LLMs exhibit stronger mathematical reasoning}, reflected in lower Miss Rates, higher Win Rates, and higher PRM Accuracy on average. Prior work has shown that LLMs can produce the correct final answer despite flawed or incomplete reasoning \citep{mondorf2024accuracyevaluatingreasoningbehavior}. Our findings demonstrate that access to external tools often amplifies this discrepancy. 

Interestingly, the highest PRM scores are achieved by non-thinking GPT-4.1 models compared to their stronger ``thinking'' variants. This is inline with recent findings that step-level reward models struggle to reliably assess long and complex reasoning chains, often conflating fluency with correctness, becoming miscalibrated on stronger models, and generalizing poorly to the longer, self-correcting traces produced by advanced reasoning models~\citep{BambaRewardUA, lee2025rethinkingrewardmodelsmultidomain}.

Finally, we posit that the reasoning gap between Base-LLM and TaLM appears moderate in aggregate because \task{} is partially masked by three factors: (1) our filtering for \task-prone problems, while targeted, is not perfect; (2) existing metrics have limited sensitivity to \task{} hallucinations; and (3) tool use yields only modest accuracy gains for most LLMs. For subsequent analyses, we focus only on solutions with correct final answers.

\begin{figure*}
    \centering
    \includegraphics[width=0.98\linewidth]{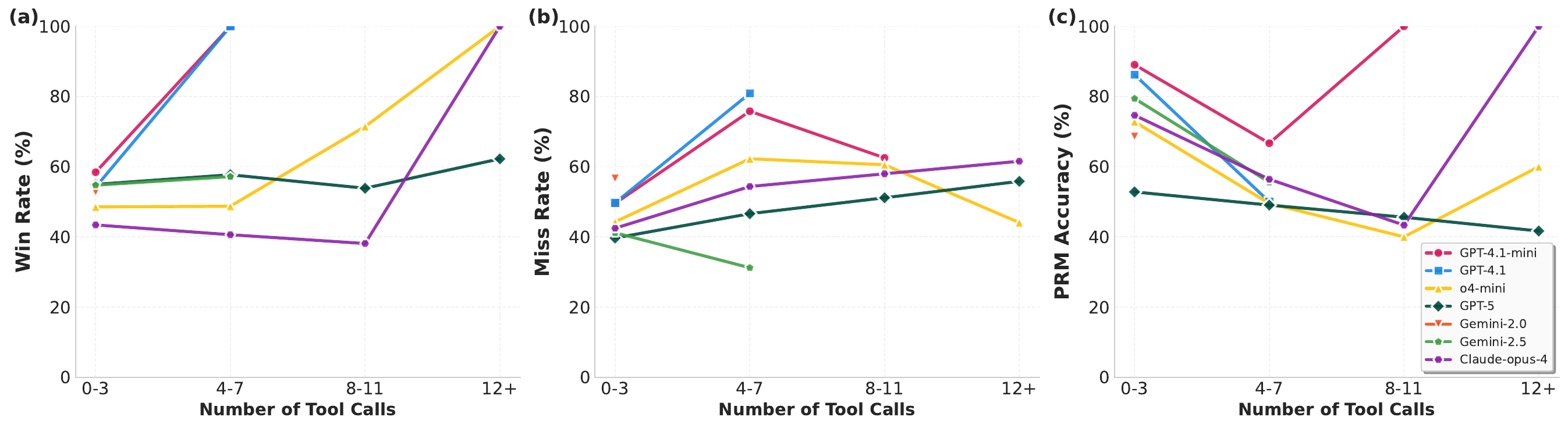}
    \caption{Impact of tool-call frequency on TaLM reasoning. Across tool-call bins (0–3 to 12+), higher tool-call frequency is associated with (a) increased Base-over-TaLM Win Rate, (b) higher Miss Rate, and (c) lower PRM Accuracy, indicating a widening reasoning gap as reliance on tools grows.
    }
    \label{fig:metrics_categorized}
    \postspace
\end{figure*}

\subsection{Base-TaLM Reasoning Gap Widens with Number of Tool Calls}
\label{sec:tool_call_and_tim}

We examine how reasoning behavior varies with tool-call frequency, a key factor underlying \task: as tool reliance grows, models increasingly substitute computation for reasoning. We test whether higher number of calls amplify the Base–TaLM gap when final answers are correct. We group problems into bins based on the number of tool calls in their TaLM-generated solutions: \{0--3, 4--7, 8--11, 12+\}. Figure~\ref{fig:metrics_categorized} shows three consistent trends across most models.
First, the Base model’s \emph{Win Rate} against the TaLM increases with tool-call frequency, indicating degraded reasoning under heavier tool reliance. Second, the \emph{Miss Rate} generally rises with increased tool calls, reflecting greater divergence from reference reasoning. Third, \emph{PRM Accuracy} typically declines as tool calls increase, suggesting that longer, tool-heavy trajectories accumulate step-level errors. Occasional reversals in the 12+ bin are due to limited sample sizes. 
We further verify that this effect is not explained by the syntactic or structural complexity of generated code; detailed analysis shows no correlation between code complexity and \task{} severity (Appendix~\ref{appendix:code_complexity}).

\subsection{Tool Use Induces Global Reasoning Errors}
\label{sec:error_type_analysis}
In this section, we analyze how reasoning error types shift under tool use by comparing TaLM solutions with those of Base-LLMs. 
Using the error taxonomy adapted from \citet{petrov2025proofbluffevaluatingllms}, we prompt our LLM judge (GPT-5) to rate each error type on a 1--5 scale, or output \texttt{None} if no error is detected (prompt in Appendix~\ref{appendix:prevalent_error_annotation}). 

\begin{figure}
    \centering
    \includegraphics[width=0.98\linewidth]{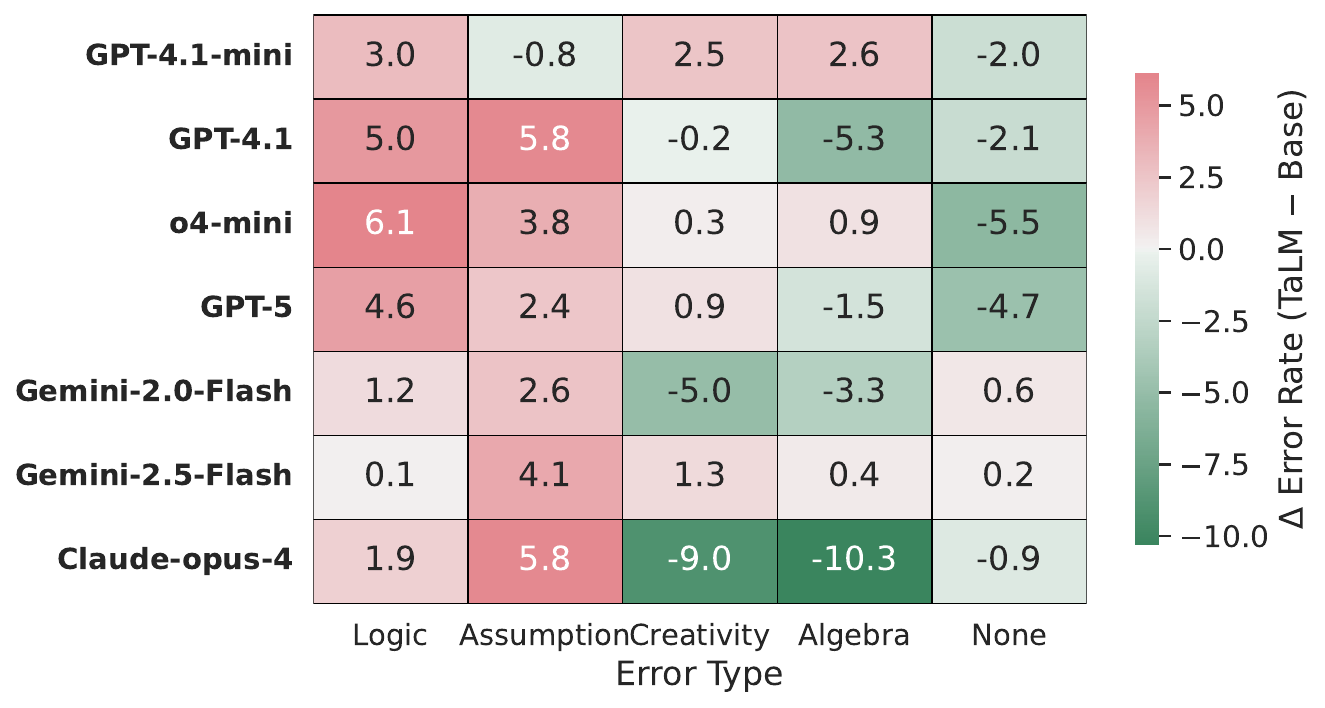}
    \caption{Change in reasoning error rates after tool use (\(\Delta = \text{TaLM} - \text{Base}\)). Positive values indicate higher error frequency with Code Interpreter access.}
    \label{fig:error_rates_delta}
    \postspace
\end{figure}

Across models, tool use induces a consistent shift in error patterns (Figure~\ref{fig:error_rates_delta}). Logic, assumption, and creativity errors increase in nearly all cases, indicating more unjustified inferences and reasoning shortcuts under tool reliance. In contrast, algebraic and arithmetic errors decrease as computation is delegated to the Code Interpreter, while the rate of error-free solutions also drops. These results show that \task{} is not merely about missing steps, but reflects a fundamental change in how models reason under tool use.

\subsection{Thinking Models Use Code More Often} 
\label{sec:tool_invocation_rate}

\begin{figure}
    \centering
    \includegraphics[width=0.98\linewidth]{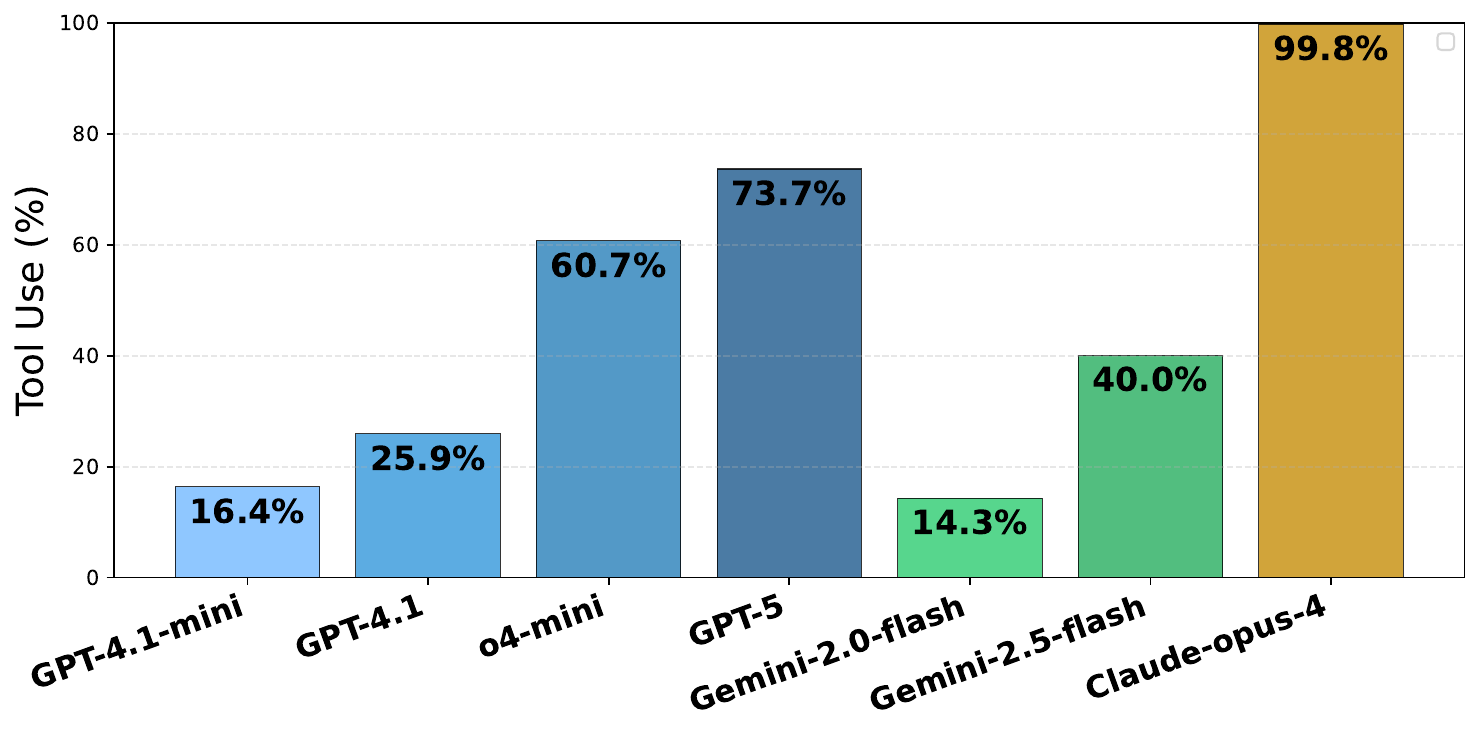}
    \caption{Tool use rate across TaLMs. Thinking TaLMs (GPT-5, o4-mini, Gemini-2.5-Flash) invoke code on \textasciitilde50\% more problems than non-thinking models.
    }
    \vspace{-3mm}
    \label{fig:tool_use_rate}
\end{figure}

Having shown that more tool calls intensify \task{}, we now quantify how often each model invokes the Code Interpreter tool on math problems, providing a ``dose'' measure of \task{} risk. We investigate tool use rates across TaLMs on \dataset, which was curated to encourage code use. Figure~\ref{fig:tool_use_rate} shows the percentage of problems on which each TaLM invoked the Code Interpreter tool. Thinking models exhibit substantially higher tool usage than non-thinking models: Claude-opus-4 achieves the highest rate at 99.8\%, followed by GPT-5 at 73.7\%, while non-thinking models show more modest usage.
\textbf{On average, thinking models invoke code on 49.7\% more problems.} Given the established relationship between tool-call frequency and \task{} severity (Section~\ref{sec:tool_call_and_tim}), this heavier reliance suggests that thinking models face higher \task{} risk.

\subsection{\task{} in Over Half of High-Risk Solutions}
\label{sec:quality_analysis}
We conduct a targeted manual evaluation to validate \task{} qualitatively and identify recurring linguistic cues that precede it. We focus on high-risk TaLM solutions that satisfy three criteria: (1) a correct final answer, (2) Base-LLM solution is preferred over the TaLM solution in Win Rate judgment, and (3) PRM flags errors in the TaLM’s reasoning trace. From this filtered set, we select the top 10 samples per model ranked by Miss Rate to assess \task{} presence and characteristic precursor phrases.


Across models, \textbf{\task{} appears in 54.3\% of high-risk cases}, demonstrating strong alignment between our automated metrics. Incidence is higher for models with greater tool use (Figure~\ref{fig:tool_use_rate}), such as Claude-Opus-4 and o4-mini, and lower for Gemini models. Less capable models often display explicit precursor cues (e.g., ``one numerically finds,'' ``systematic checks show,'' or ``let's verify programmatically.''), whereas stronger models like GPT-5 silently substitute code outputs for derivations (more details and a list of precursor phrases in Appendix~\ref{appendix:common-precursor-phrases}). These cues are indicative rather than definitive and should be interpreted alongside quantitative metrics.

\section{Mitigating \task{} Hallucination}
\label{sec:talm_tuning}
Our evaluation results reveal widespread \task{} hallucinations, where models substitute tool outputs for mathematical reasoning. In this section, we propose two complementary mitigation strategies that encourage models to use the Code Interpreter as an \textit{assistant to reasoning} rather than a substitute.


\subsection{Prompting-Based Mitigation}
\label{sec:prompting_mitigation}

We design a lightweight prompting strategy that self-instructs the model to treat tool outputs as reasoning aids. Specifically, after each problem statement, we inject the following instruction:

\begin{quote}
\vspace{-0.5mm}
\textit{``We should treat code snippets and their execution results only as helpful hints, and derive the solution through mathematical reasoning.''}
\vspace{-0.5mm}
\end{quote}

\noindent
This single-sentence intervention promotes mathematical thinking over computational shortcuts.


\subsection{Alignment-Based Mitigation}
\label{sec:alignment_tuning}

We view \task{} primarily as a behavioral misalignment issue: under tool use, the model tends to treat Code Interpreter as a search mechanism that substitutes for derivation, rather than as an assistant that supports reasoning. This suggests that mitigation should target \emph{how} the model reasons with tool outputs, not merely whether it reaches the correct answer. We therefore adopt Direct Preference Optimization (DPO; \citep{10.5555/3666122.3668460}), which is well-suited for aligning relative preferences between alternative reasoning trajectories under the same problem and tool context. Due to resource constraints, we focus on fine-tuning a single model (GPT-4.1) to test this approach.



\subsubsection{Preference Data Creation}
\label{sec:data_collection}

We construct a preference dataset from the training split of \dataset{}. Because \task{} is a reasoning failure rather than an answer-correctness failure, we explicitly control for correctness in our preference construction: both chosen and rejected samples are required to preserve the same intermediate and final results. Thus, the preference signal cannot be explained by answer correctness, but instead isolates differences in reasoning behavior under tool use. Accordingly, for each problem, we construct a \textit{chosen-rejected} pair, where the chosen solution exhibits stronger reasoning under tool use and the rejected solution exhibits \task{} characteristics.

\noindent
\textbf{Chosen samples:} We generate chosen solutions by applying the prompting strategy from Section~\ref{sec:prompting_mitigation} to GPT-4.1, which encourages mathematical reasoning under tool use.

\noindent
\textbf{Rejected samples:} We generate rejected examples through controlled degradation: given a problem and its corresponding chosen solution,
we prompt the same model to rewrite a solution span with excessive reliance on tool outputs, producing a coherent but tool-dependent solution that omits or abbreviates intermediate reasoning steps.
This process creates naturalistic rejected samples with \task{} hallucination 
(more details in Appendix~\ref{appendix:preference-optimization-framework}). 

\subsection{Results}



\begin{table}[t]
\centering
\resizebox{0.49\textwidth}{!}{
\begin{tabular}{l c c c c}
\toprule
\textbf{Variant} & \textbf{Final Acc.} ($\uparrow$) & \textbf{Miss Rate} ($\downarrow$) & \textbf{Win Rate vs. TaLM} ($\uparrow$) & \textbf{PRM Acc.} ($\uparrow$) \\
\midrule
 Base & {\large 24.6} & {\large 48.1} & {\large 54.4} & \textbf{\large 88.6} \\
 TaLM & {\large 27.0} & {\large 49.9} & {\large --} & {\large 85.9} \\
 \rowcolor{blue!6}\textbf{TaLM + Prompting} & {\large 25.1} & {\large 49.4} & {\large 52.7} & {\large 82.9} \\
 \rowcolor{green!8}\textbf{TaLM + DPO} & \textbf{\large 27.6} & \textbf{\large 46.6} & \textbf{\large 58.2} & {\large 83.3} \\
\bottomrule
\end{tabular}
}
\caption{Impact of mitigation strategies on tool-augmented GPT-4.1. Prompting reduces \task{} without training but at the cost of accuracy, while DPO improves both accuracy and reasoning quality.}

\label{tab:mitigation_results}
\end{table}

Table~\ref{tab:mitigation_results} presents mitigation results on our evaluation benchmark. Both mitigation strategies reduce \task{} hallucination compared to the vanilla TaLM, but with distinct trade-offs.

\noindent
\textbf{Prompting intervention.} Zero-shot prompting substantially improves reasoning without model retraining, improving Win Rate from $45.6\%$ to $52.7\%$ and slightly reducing Miss Rate ($49.9\%$ to $49.4\%$). This improvement comes at a cost to final-answer accuracy, which drops from $27.0\%$ to $25.1\%$, approaching the Base-LLM's $24.6\%$.

\noindent
\textbf{DPO alignment.}  The fine-tuned model achieves the strongest overall trade-off: it attains the highest final-answer accuracy ($27.6\%$, +0.6 points over the vanilla TaLM), the lowest Miss Rate ($46.6\%$), and is preferred over the vanilla TaLM in $58.2\%$ of pairwise comparisons. However, its PRM accuracy ($83.3\%$) remains below that of the Base-LLM ($88.6\%$), suggesting that while DPO mitigates global reasoning failures, it does not fully recover step-level correctness. The largest gains appear on the reasoning-sensitive metrics, especially Miss Rate and Win Rate, while final-answer accuracy improves only modestly. These results suggest that DPO primarily acts as a behavioral alignment mechanism rather than a capability boost.


\textbf{DPO alignment.} The fine-tuned model achieves the strongest overall performance: highest final-answer accuracy ($27.6\%$, +0.6\% over vanilla TaLM), highest Win Rate ($58.2\%$ vs. $54.4\%$ for the Base-LLM), and lowest Miss Rate ($46.6\%$). However, PRM accuracy ($83.3\%$) remains below that of the Base-LLM ($88.6\%$), suggesting that while DPO mitigates global reasoning failures, it does not fully restore step-level correctness. Notably, the largest gains appear in the reasoning-sensitive metrics, especially Win Rate and Miss Rate, while final-answer accuracy improves only modestly. This pattern is consistent with a behavioral alignment effect: DPO primarily changes how the model uses tool outputs during reasoning, rather than simply improving task capability.




\section{Related Work}

Tool-augmented reasoning extends LLMs with external programmatic interfaces to overcome limitations of parametric knowledge. Early work, such as Toolformer~\citep{toolformer}, showed that models can learn when and how to invoke tools, while later surveys~\citep{wang2024toolsanywaysurveylanguage, gou2024tora} define LLM-used tools and provide broad overviews of tool-augmented LMs. This has led to the development of diverse tools, such as web search~\citep{asai2024selfrag, 10.5555/3666122.3668004}, code interpreters~\citep{POTprompting, gao2022pal}, and domain-specific APIs~\citep{Qin2023ToolLLMFL}.

\paragraph{Hallucinations in TaLMs.}
Despite these advances, TaLMs exhibit persistent failure modes. Prior work has studied \textit{tool hallucinations}, where models misuse or select incorrect tools~\citep{xu2025reducingtoolhallucinationreliability}, as well as factual errors arising from conflicts between parametric and retrieved knowledge~\citep{sun2025redeepdetectinghallucinationretrievalaugmented} or from inaccurate retrieved content~\citep{magesh2024hallucinationfreeassessingreliabilityleading}. Other studies highlight attribution failures in retrieval-augmented generation~\citep{gao-etal-2023-enabling, liu-etal-2023-evaluating} and cognitive offloading, where unrestricted tool access suppresses internal reasoning~\citep{qian-etal-2025-smart, wang2025actingreasoningmoreteaching}.  
In contrast, we show that \emph{even when tools are correctly invoked and executed}, TaLMs exhibit Tool-Induced Myopia, which causes them to substitute reasoning with computation.

\paragraph{Evaluating Mathematical Reasoning.}
Evaluation of mathematical reasoning has traditionally relied on final-answer accuracy using benchmarks such as GSM8K~\citep{cobbe2021gsm8k} and MATH~\citep{hendrycks2021math}. However, outcome-based metrics fail to capture reasoning behavior, especially as frontier models approach saturation~\citep{mondorf2024accuracyevaluatingreasoningbehavior}. Recent work has proposed process-level evaluation ~\citep{lightman2023letsverifystepstep, zheng-etal-2025-processbench}, chain-of-thought scoring~\citep{xiareasoneval, li-etal-2025-fg}, and model-based judges ~\citep{mathcheck, xu2025reducingtoolhallucinationreliability}. However, even with competition-level benchmarks~\citep{AIME20242025, he-etal-2024-olympiadbench, gao2024omnimathuniversalolympiadlevel}, existing evaluations remain largely one-dimensional and fail to distinguish robust reasoning from tool-driven shortcuts. 
We address this gap with our benchmark and multi-dimensional evaluation suite explicitly designed to surface \task{}.

\paragraph{Mitigating TaLM Failures.}
Prior mitigation strategies focus on prompting-based self-correction~\citep{shinn2023reflexion}, alignment techniques targeting tool selection and invocation errors~\citep{qintoollearning, patilGorilla, ross-etal-2025-when2call, xu2025reducingtoolhallucinationreliability}, and hybrid approaches that combine parametric and non-parametric reasoning~\citep{asai2024selfrag}. While effective for tool misuse, these approaches do not address reasoning degradation under correct tool use. Our work introduces interventions that explicitly reward comprehensive analytical reasoning over premature tool reliance, providing the first systematic approach to mitigating \task.

\section{Conclusion}
Investigating tool-augmented reasoning across competition-level math, we show that even with correct tool use, TaLMs exhibit \textbf{Tool-Induced Myopia (\task)}, substituting computation for mathematical reasoning. To study this, we introduced \dataset{} and a four-dimensional evaluation suite that reveals a consistent pattern across seven frontier models: TaLMs improve final-answer accuracy but produce weaker reasoning than Base-LLMs. Error analysis shows fewer arithmetic errors but more logical and assumption errors, especially with increased tool use. Finally, we introduced two mitigation strategies: prompting and DPO-based preference optimization, which reduce \task{} hallucination and improve trustworthy reasoning.


\section{Limitations}

While our study provides the first characterization of Tool-Induced Myopia (\task{}), it also has several limitations:
First, to ensure precise control over tool invocation and execution, we restrict our analysis to a single tool, the Code Interpreter. This choice eliminates confounds such as API failures or retrieval noise, but it limits the generality of our findings to other tool types (e.g., search, retrieval, or various APIs). Future work should extend this framework to a broader range of tools and interaction settings.
Second, our manual investigation of \task{} occurrence and linguistic precursors is limited in scope. While qualitative inspection confirms strong alignment with our automated metrics, a larger-scale human evaluation, covering more models, domains, and annotators, would strengthen the robustness and generalizability of these observations.
Finally, due to computational and resource constraints, our mitigation experiments focus exclusively on the GPT-4.1. Although this choice provides a representative case for high-end tool-augmented reasoning, reproducing the mitigation analyses on additional LLMs remains an important direction for validating consistency and scalability.

\section{Ethics Statement}

This work studies the reasoning behavior of tool-augmented large language models using publicly available mathematical problem datasets and model-generated outputs. No human subjects were involved in data collection, and all qualitative annotations were performed by the authors on model outputs only.

We use proprietary language models accessed via official APIs and follow the providers’ usage policies. Our analysis focuses on identifying and mitigating reasoning failures in tool-augmented models, with the goal of improving reliability and trustworthiness in deployed systems. We do not release any personally identifiable information, and all data used or released are either publicly available or derived from synthetic model outputs.

AI assistants, ChatGPT and Claude models, were used for writing support and minor editing. All scientific contributions, experimental design, analysis, and conclusions are the responsibility of the authors.

\bibliography{custom}

@inproceedings{yu-ji-2024-information,
    title = "Information Association for Language Model Updating by Mitigating {LM}-Logical Discrepancy",
    author = "Yu, Pengfei  and
      Ji, Heng",
    editor = "Barak, Libby  and
      Alikhani, Malihe",
    booktitle = "Proceedings of the 28th Conference on Computational Natural Language Learning",
    month = nov,
    year = "2024",
    address = "Miami, FL, USA",
    publisher = "Association for Computational Linguistics",
    url = "https://aclanthology.org/2024.conll-1.10/",
    doi = "10.18653/v1/2024.conll-1.10",
    pages = "117--129"
}

@misc{wang2025actingreasoningmoreteaching,
      title={Acting Less is Reasoning More! Teaching Model to Act Efficiently}, 
      author={Hongru Wang and Cheng Qian and Wanjun Zhong and Xiusi Chen and Jiahao Qiu and Shijue Huang and Bowen Jin and Mengdi Wang and Kam-Fai Wong and Heng Ji},
      year={2025},
      eprint={2504.14870},
      archivePrefix={arXiv},
      primaryClass={cs.AI},
      url={https://arxiv.org/abs/2504.14870}, 
}

@inproceedings{qian-etal-2024-tell,
    title = "Tell Me More! Towards Implicit User Intention Understanding of Language Model Driven Agents",
    author = "Qian, Cheng  and
      He, Bingxiang  and
      Zhuang, Zhong  and
      Deng, Jia  and
      Qin, Yujia  and
      Cong, Xin  and
      Zhang, Zhong  and
      Zhou, Jie  and
      Lin, Yankai  and
      Liu, Zhiyuan  and
      Sun, Maosong",
    editor = "Ku, Lun-Wei  and
      Martins, Andre  and
      Srikumar, Vivek",
    booktitle = "Proceedings of the 62nd Annual Meeting of the Association for Computational Linguistics (Volume 1: Long Papers)",
    month = aug,
    year = "2024",
    address = "Bangkok, Thailand",
    publisher = "Association for Computational Linguistics",
    url = "https://aclanthology.org/2024.acl-long.61/",
    doi = "10.18653/v1/2024.acl-long.61",
    pages = "1088--1113"
}

@inproceedings{lu-etal-2023-survey,
    title = "A Survey of Deep Learning for Mathematical Reasoning",
    author = "Lu, Pan  and
      Qiu, Liang  and
      Yu, Wenhao  and
      Welleck, Sean  and
      Chang, Kai-Wei",
    editor = "Rogers, Anna  and
      Boyd-Graber, Jordan  and
      Okazaki, Naoaki",
    booktitle = "Proceedings of the 61st Annual Meeting of the Association for Computational Linguistics (Volume 1: Long Papers)",
    month = jul,
    year = "2023",
    address = "Toronto, Canada",
    publisher = "Association for Computational Linguistics",
    url = "https://aclanthology.org/2023.acl-long.817/",
    doi = "10.18653/v1/2023.acl-long.817",
    pages = "14605--14631"
}

@inproceedings{gou2024tora,
title={To{RA}: A Tool-Integrated Reasoning Agent for Mathematical Problem Solving},
author={Zhibin Gou and Zhihong Shao and Yeyun Gong and yelong shen and Yujiu Yang and Minlie Huang and Nan Duan and Weizhu Chen},
booktitle={The Twelfth International Conference on Learning Representations},
year={2024},
url={https://openreview.net/forum?id=Ep0TtjVoap}
}

@misc{wang2024toolsanywaysurveylanguage,
      title={What Are Tools Anyway? A Survey from the Language Model Perspective}, 
      author={Zhiruo Wang and Zhoujun Cheng and Hao Zhu and Daniel Fried and Graham Neubig},
      year={2024},
      eprint={2403.15452},
      archivePrefix={arXiv},
      primaryClass={cs.CL},
      url={https://arxiv.org/abs/2403.15452}, 
}

@inproceedings{toolformer,
author = {Schick, Timo and Dwivedi-Yu, Jane and Dess\'{\i}, Roberto and Raileanu, Roberta and Lomeli, Maria and Hambro, Eric and Zettlemoyer, Luke and Cancedda, Nicola and Scialom, Thomas},
title = {Toolformer: language models can teach themselves to use tools},
year = {2023},
publisher = {Curran Associates Inc.},
address = {Red Hook, NY, USA},
booktitle = {Proceedings of the 37th International Conference on Neural Information Processing Systems},
articleno = {2997},
numpages = {13},
location = {New Orleans, LA, USA},
series = {NIPS '23}
}

@article{Qin2023ToolLLMFL,
  title={ToolLLM: Facilitating Large Language Models to Master 16000+ Real-world APIs},
  author={Yujia Qin and Shi Liang and Yining Ye and Kunlun Zhu and Lan Yan and Ya-Ting Lu and Yankai Lin and Xin Cong and Xiangru Tang and Bill Qian and Sihan Zhao and Runchu Tian and Ruobing Xie and Jie Zhou and Marc H. Gerstein and Dahai Li and Zhiyuan Liu and Maosong Sun},
  journal={ArXiv},
  year={2023},
  volume={abs/2307.16789},
  url={https://api.semanticscholar.org/CorpusID:260334759}
}

@article{POTprompting,
  title = {Program of Thoughts Prompting: Disentangling Computation from Reasoning for Numerical Reasoning Tasks},
  author = {Wenhu Chen and Xueguang Ma and Xinyi Wang and William W. Cohen},
  journal={Transactions on Machine Learning Research},
  year = {2023},
}

@article{gao2022pal,
  title={PAL: Program-aided Language Models},
  author={Gao, Luyu and Madaan, Aman and Zhou, Shuyan and Alon, Uri and Liu, Pengfei and Yang, Yiming and Callan, Jamie and Neubig, Graham},
  journal={Proceedings of the 40th International Conference on Machine Learning},
  year={2023}
}

@inproceedings{10.5555/3666122.3668004,
author = {Lu, Pan and Peng, Baolin and Cheng, Hao and Galley, Michel and Chang, Kai-Wei and Wu, Ying Nian and Zhu, Song-Chun and Gao, Jianfeng},
title = {Chameleon: plug-and-play compositional reasoning with large language models},
year = {2023},
publisher = {Curran Associates Inc.},
address = {Red Hook, NY, USA},
booktitle = {Proceedings of the 37th International Conference on Neural Information Processing Systems},
articleno = {1882},
numpages = {32},
location = {New Orleans, LA, USA},
series = {NIPS '23}
}

@misc{xu2025reducingtoolhallucinationreliability,
      title={Reducing Tool Hallucination via Reliability Alignment}, 
      author={Hongshen Xu and Zichen Zhu and Lei Pan and Zihan Wang and Su Zhu and Da Ma and Ruisheng Cao and Lu Chen and Kai Yu},
      year={2025},
      eprint={2412.04141},
      publisher={Proceedings of the 42nd International Conference on Machine Learning},
      primaryClass={cs.CL},
      url={https://arxiv.org/abs/2412.04141}, 
}

@article{asai2024selfrag,
      author    = {Asai, Akari and Wu, Zeqiu and Wang, Yizhong and Sil, Avirup and Hajishirzi, Hannaneh},
      title     = {{Self-RAG}: Learning to Retrieve, Generate, and Critique through Self-Reflection},
      year      = {2024},
     journal    = {Proceedings of the International Conference on Learning Representations (ICLR) 2024},
     url        = {https://arxiv.org/abs/2310.11511}
    }

@inproceedings{10.5555/3737916.3741936,
author = {Patil, Shishir G. and Zhang, Tianjun and Wang, Xin and Gonzalez, Joseph E.},
title = {Gorilla: large language model connected with massive APIs},
year = {2025},
isbn = {9798331314385},
publisher = {Curran Associates Inc.},
address = {Red Hook, NY, USA},
booktitle = {Proceedings of the 38th International Conference on Neural Information Processing Systems},
articleno = {4020},
numpages = {22},
location = {Vancouver, BC, Canada},
series = {NIPS '24}
}

@misc{sun2025redeepdetectinghallucinationretrievalaugmented,
      title={ReDeEP: Detecting Hallucination in Retrieval-Augmented Generation via Mechanistic Interpretability}, 
      author={Zhongxiang Sun and Xiaoxue Zang and Kai Zheng and Yang Song and Jun Xu and Xiao Zhang and Weijie Yu and Yang Song and Han Li},
      year={2025},
      eprint={2410.11414},
      archivePrefix={arXiv},
      primaryClass={cs.CL},
      url={https://arxiv.org/abs/2410.11414}, 
}

@misc{magesh2024hallucinationfreeassessingreliabilityleading,
      title={Hallucination-Free? Assessing the Reliability of Leading AI Legal Research Tools}, 
      author={Varun Magesh and Faiz Surani and Matthew Dahl and Mirac Suzgun and Christopher D. Manning and Daniel E. Ho},
      year={2024},
      eprint={2405.20362},
      archivePrefix={arXiv},
      primaryClass={cs.CL},
      url={https://arxiv.org/abs/2405.20362}, 
}

@inproceedings{qian-etal-2025-smart,
    title = "{SMART}: Self-Aware Agent for Tool Overuse Mitigation",
    author = {Qian, Cheng  and
      Acikgoz, Emre Can  and
      Wang, Hongru  and
      Chen, Xiusi  and
      Sil, Avirup  and
      Hakkani-T{\"u}r, Dilek  and
      Tur, Gokhan  and
      Ji, Heng},
    editor = "Che, Wanxiang  and
      Nabende, Joyce  and
      Shutova, Ekaterina  and
      Pilehvar, Mohammad Taher",
    booktitle = "Findings of the Association for Computational Linguistics: ACL 2025",
    month = jul,
    year = "2025",
    address = "Vienna, Austria",
    publisher = "Association for Computational Linguistics",
    url = "https://aclanthology.org/2025.findings-acl.239/",
    doi = "10.18653/v1/2025.findings-acl.239",
    pages = "4604--4621",
    ISBN = "979-8-89176-256-5"
}

@inproceedings{10.5555/3666122.3668460,
author = {Rafailov, Rafael and Sharma, Archit and Mitchell, Eric and Ermon, Stefano and Manning, Christopher D. and Finn, Chelsea},
title = {Direct preference optimization: your language model is secretly a reward model},
year = {2023},
publisher = {Curran Associates Inc.},
address = {Red Hook, NY, USA},
booktitle = {Proceedings of the 37th International Conference on Neural Information Processing Systems},
articleno = {2338},
numpages = {14},
location = {New Orleans, LA, USA},
series = {NIPS '23}
}

@misc{gu2025surveyllmasajudge,
      title={A Survey on LLM-as-a-Judge}, 
      author={Jiawei Gu and Xuhui Jiang and Zhichao Shi and Hexiang Tan and Xuehao Zhai and Chengjin Xu and Wei Li and Yinghan Shen and Shengjie Ma and Honghao Liu and Saizhuo Wang and Kun Zhang and Yuanzhuo Wang and Wen Gao and Lionel Ni and Jian Guo},
      year={2025},
      eprint={2411.15594},
      archivePrefix={arXiv},
      primaryClass={cs.CL},
      url={https://arxiv.org/abs/2411.15594}, 
}

@misc{AIME20242025,
  title        = {AIME I and AIME II Problems and Solutions},
  howpublished = {\url{https://artofproblemsolving.com/wiki/index.php/AIME_Problems_and_Solutions}},
  year         = {2024-2025},
}

@inproceedings{huang2024olympicarena,
 author = {Huang, Zhen and Wang, Zengzhi and Xia, Shijie and Li, Xuefeng and Zou, Haoyang and Xu, Ruijie and Fan, Run-Ze and Ye, Lyumanshan and Chern, Ethan and Ye, Yixin and Zhang, Yikai and Yang, Yuqing and Wu, Ting and Wang, Binjie and Sun, Shichao and Xiao, Yang and Li, Yiyuan and Zhou, Fan and Chern, Steffi and Qin, Yiwei and Ma, Yan and Su, Jiadi and Liu, Yixiu and Zheng, Yuxiang and Zhang, Shaoting and Lin, Dahua and Qiao, Yu and Liu, Pengfei},
 booktitle = {Advances in Neural Information Processing Systems},
 editor = {A. Globerson and L. Mackey and D. Belgrave and A. Fan and U. Paquet and J. Tomczak and C. Zhang},
 pages = {19209--19253},
 publisher = {Curran Associates, Inc.},
 title = {OlympicArena: Benchmarking Multi-discipline Cognitive Reasoning for Superintelligent AI},
 url = {https://proceedings.neurips.cc/paper_files/paper/2024/file/222d2eaf24cf8259a35d6c7130d31425-Paper-Datasets_and_Benchmarks_Track.pdf},
 volume = {37},
 year = {2024}
}

@inproceedings{he-etal-2024-olympiadbench,
    title = "{O}lympiad{B}ench: A Challenging Benchmark for Promoting {AGI} with Olympiad-Level Bilingual Multimodal Scientific Problems",
    author = "He, Chaoqun  and
      Luo, Renjie  and
      Bai, Yuzhuo  and
      Hu, Shengding  and
      Thai, Zhen  and
      Shen, Junhao  and
      Hu, Jinyi  and
      Han, Xu  and
      Huang, Yujie  and
      Zhang, Yuxiang  and
      Liu, Jie  and
      Qi, Lei  and
      Liu, Zhiyuan  and
      Sun, Maosong",
    editor = "Ku, Lun-Wei  and
      Martins, Andre  and
      Srikumar, Vivek",
    booktitle = "Proceedings of the 62nd Annual Meeting of the Association for Computational Linguistics (Volume 1: Long Papers)",
    month = aug,
    year = "2024",
    address = "Bangkok, Thailand",
    publisher = "Association for Computational Linguistics",
    url = "https://aclanthology.org/2024.acl-long.211/",
    doi = "10.18653/v1/2024.acl-long.211",
    pages = "3828--3850"
}

@misc{gao2024omnimathuniversalolympiadlevel,
      title={Omni-MATH: A Universal Olympiad Level Mathematic Benchmark For Large Language Models}, 
      author={Bofei Gao and Feifan Song and Zhe Yang and Zefan Cai and Yibo Miao and Qingxiu Dong and Lei Li and Chenghao Ma and Liang Chen and Runxin Xu and Zhengyang Tang and Benyou Wang and Daoguang Zan and Shanghaoran Quan and Ge Zhang and Lei Sha and Yichang Zhang and Xuancheng Ren and Tianyu Liu and Baobao Chang},
      year={2024},
      eprint={2410.07985},
      archivePrefix={arXiv},
      primaryClass={cs.CL},
      url={https://arxiv.org/abs/2410.07985}, 
}

@misc{petrov2025proofbluffevaluatingllms,
      title={Proof or Bluff? Evaluating LLMs on 2025 USA Math Olympiad}, 
      author={Ivo Petrov and Jasper Dekoninck and Lyuben Baltadzhiev and Maria Drencheva and Kristian Minchev and Mislav Balunović and Nikola Jovanović and Martin Vechev},
      year={2025},
      eprint={2503.21934},
      archivePrefix={arXiv},
      primaryClass={cs.CL},
      url={https://arxiv.org/abs/2503.21934}, 
}

@inproceedings{zheng-etal-2025-processbench,
    title = "{P}rocess{B}ench: Identifying Process Errors in Mathematical Reasoning",
    author = "Zheng, Chujie  and
      Zhang, Zhenru  and
      Zhang, Beichen  and
      Lin, Runji  and
      Lu, Keming  and
      Yu, Bowen  and
      Liu, Dayiheng  and
      Zhou, Jingren  and
      Lin, Junyang",
    editor = "Che, Wanxiang  and
      Nabende, Joyce  and
      Shutova, Ekaterina  and
      Pilehvar, Mohammad Taher",
    booktitle = "Proceedings of the 63rd Annual Meeting of the Association for Computational Linguistics (Volume 1: Long Papers)",
    month = jul,
    year = "2025",
    address = "Vienna, Austria",
    publisher = "Association for Computational Linguistics",
    url = "https://aclanthology.org/2025.acl-long.50/",
    doi = "10.18653/v1/2025.acl-long.50",
    pages = "1009--1024",
    ISBN = "979-8-89176-251-0"
}

@article{10.1145/3773985,
author = {Liu, WenTao and Hu, Hanglei and Zhou, Jie and Ding, Yuyang and Li, Junsong and Zeng, Jiayi and He, MengLiang and Chen, Qin and Jiang, Bo and Zhou, Aimin and He, Liang},
title = {Mathematical Language Models: A Survey},
year = {2025},
publisher = {Association for Computing Machinery},
address = {New York, NY, USA},
issn = {0360-0300},
url = {https://doi.org/10.1145/3773985},
doi = {10.1145/3773985},
note = {Just Accepted},
journal = {ACM Comput. Surv.},
month = nov
}

@inproceedings{ahn-etal-2024-large,
    title = "Large Language Models for Mathematical Reasoning: Progresses and Challenges",
    author = "Ahn, Janice  and
      Verma, Rishu  and
      Lou, Renze  and
      Liu, Di  and
      Zhang, Rui  and
      Yin, Wenpeng",
    editor = "Falk, Neele  and
      Papi, Sara  and
      Zhang, Mike",
    booktitle = "Proceedings of the 18th Conference of the European Chapter of the Association for Computational Linguistics: Student Research Workshop",
    month = mar,
    year = "2024",
    address = "St. Julian{'}s, Malta",
    publisher = "Association for Computational Linguistics",
    url = "https://aclanthology.org/2024.eacl-srw.17/",
    doi = "10.18653/v1/2024.eacl-srw.17",
    pages = "225--237"
}

@misc{lightman2023letsverifystepstep,
      title={Let's Verify Step by Step}, 
      author={Hunter Lightman and Vineet Kosaraju and Yura Burda and Harri Edwards and Bowen Baker and Teddy Lee and Jan Leike and John Schulman and Ilya Sutskever and Karl Cobbe},
      year={2023},
      eprint={2305.20050},
      archivePrefix={arXiv},
      primaryClass={cs.LG},
      url={https://arxiv.org/abs/2305.20050}, 
}

@inproceedings{mondorf2024accuracyevaluatingreasoningbehavior,
title={Beyond Accuracy: Evaluating the Reasoning Behavior of Large Language Models -- A Survey}, 
author={Philipp Mondorf and Barbara Plank},
booktitle={First Conference on Language Modeling},
year={2024},
url={https://arxiv.org/abs/2404.01869}, 
}

@inproceedings{yee2024faithful,
title={Faithful and Unfaithful Error Recovery in Chain of Thought},
author={Evelyn Yee and Alice Li and Chenyu Tang and Yeon Ho Jung and Ramamohan Paturi and Leon Bergen},
booktitle={First Conference on Language Modeling},
year={2024},
url={https://openreview.net/forum?id=IPZ28ZqD4I}
}

@inproceedings{10.5555/3666122.3668142,
author = {Zheng, Lianmin and Chiang, Wei-Lin and Sheng, Ying and Zhuang, Siyuan and Wu, Zhanghao and Zhuang, Yonghao and Lin, Zi and Li, Zhuohan and Li, Dacheng and Xing, Eric P. and Zhang, Hao and Gonzalez, Joseph E. and Stoica, Ion},
title = {Judging LLM-as-a-judge with MT-bench and Chatbot Arena},
year = {2023},
publisher = {Curran Associates Inc.},
address = {Red Hook, NY, USA},
booktitle = {Proceedings of the 37th International Conference on Neural Information Processing Systems},
articleno = {2020},
numpages = {29},
location = {New Orleans, LA, USA},
series = {NIPS '23}
}

@misc{chen2025learningreasonfactuality,
      title={Learning to Reason for Factuality}, 
      author={Xilun Chen and Ilia Kulikov and Vincent-Pierre Berges and Barlas Oğuz and Rulin Shao and Gargi Ghosh and Jason Weston and Wen-tau Yih},
      year={2025},
      eprint={2508.05618},
      archivePrefix={arXiv},
      primaryClass={cs.CL},
      url={https://arxiv.org/abs/2508.05618}, 
}

@inproceedings{fatahi-bayat-etal-2025-factbench,
    title = "{F}act{B}ench: A Dynamic Benchmark for In-the-Wild Language Model Factuality Evaluation",
    author = "Fatahi Bayat, Farima  and
      Zhang, Lechen  and
      Munir, Sheza  and
      Wang, Lu",
    editor = "Che, Wanxiang  and
      Nabende, Joyce  and
      Shutova, Ekaterina  and
      Pilehvar, Mohammad Taher",
    booktitle = "Proceedings of the 63rd Annual Meeting of the Association for Computational Linguistics (Volume 1: Long Papers)",
    month = jul,
    year = "2025",
    address = "Vienna, Austria",
    publisher = "Association for Computational Linguistics",
    url = "https://aclanthology.org/2025.acl-long.1587/",
    doi = "10.18653/v1/2025.acl-long.1587",
    pages = "33090--33110",
    ISBN = "979-8-89176-251-0"
}

@inproceedings{liu-etal-2025-verifact,
    title = "{V}eri{F}act: Enhancing Long-Form Factuality Evaluation with Refined Fact Extraction and Reference Facts",
    author = "Liu, Xin  and
      Zhang, Lechen  and
      Munir, Sheza  and
      Gu, Yiyang  and
      Wang, Lu",
    editor = "Christodoulopoulos, Christos  and
      Chakraborty, Tanmoy  and
      Rose, Carolyn  and
      Peng, Violet",
    booktitle = "Proceedings of the 2025 Conference on Empirical Methods in Natural Language Processing",
    month = nov,
    year = "2025",
    address = "Suzhou, China",
    publisher = "Association for Computational Linguistics",
    url = "https://aclanthology.org/2025.emnlp-main.905/",
    pages = "17919--17936",
    ISBN = "979-8-89176-332-6"
}

@inproceedings{BambaRewardUA,
    title={Reward Under Attack: Evaluating the Sensitivity of Process Reward Models},
    author={Udbhav Bamba and Heng Yang and Rishabh Tiwari and Michael W. Mahoney and Kurt Keutzer and Amir Gholami},
    year={2025},
    url={https://api.semanticscholar.org/CorpusID:281419347}
}

@misc{lee2025rethinkingrewardmodelsmultidomain,
      title={Rethinking Reward Models for Multi-Domain Test-Time Scaling}, 
      author={Dong Bok Lee and Seanie Lee and Sangwoo Park and Minki Kang and Jinheon Baek and Dongki Kim and Dominik Wagner and Jiongdao Jin and Heejun Lee and Tobias Bocklet and Jinyu Wang and Jingjing Fu and Sung Ju Hwang and Jiang Bian and Lei Song},
      year={2025},
      eprint={2510.00492},
      archivePrefix={arXiv},
      primaryClass={cs.AI},
      url={https://arxiv.org/abs/2510.00492}, 
}

@inproceedings{gao-etal-2023-enabling,
    title = "Enabling Large Language Models to Generate Text with Citations",
    author = "Gao, Tianyu  and
      Yen, Howard  and
      Yu, Jiatong  and
      Chen, Danqi",
    editor = "Bouamor, Houda  and
      Pino, Juan  and
      Bali, Kalika",
    booktitle = "Proceedings of the 2023 Conference on Empirical Methods in Natural Language Processing",
    month = dec,
    year = "2023",
    address = "Singapore",
    publisher = "Association for Computational Linguistics",
    url = "https://aclanthology.org/2023.emnlp-main.398/",
    doi = "10.18653/v1/2023.emnlp-main.398",
    pages = "6465--6488"
}

@inproceedings{liu-etal-2023-evaluating,
    title = "Evaluating Verifiability in Generative Search Engines",
    author = "Liu, Nelson  and
      Zhang, Tianyi  and
      Liang, Percy",
    editor = "Bouamor, Houda  and
      Pino, Juan  and
      Bali, Kalika",
    booktitle = "Findings of the Association for Computational Linguistics: EMNLP 2023",
    month = dec,
    year = "2023",
    address = "Singapore",
    publisher = "Association for Computational Linguistics",
    url = "https://aclanthology.org/2023.findings-emnlp.467/",
    doi = "10.18653/v1/2023.findings-emnlp.467",
    pages = "7001--7025"
}

@misc{cobbe2021gsm8k,
      title={Training Verifiers to Solve Math Word Problems}, 
      author={Karl Cobbe and Vineet Kosaraju and Mohammad Bavarian and Mark Chen and Heewoo Jun and Lukasz Kaiser and Matthias Plappert and Jerry Tworek and Jacob Hilton and Reiichiro Nakano and Christopher Hesse and John Schulman},
      year={2021},
      eprint={2110.14168},
      archivePrefix={arXiv},
      primaryClass={cs.LG},
      url={https://arxiv.org/abs/2110.14168}, 
}

@inproceedings{hendrycks2021math,
 author = {Hendrycks, Dan and Burns, Collin and Kadavath, Saurav and Arora, Akul and Basart, Steven and Tang, Eric and Song, Dawn and Steinhardt, Jacob},
 booktitle = {Proceedings of the Neural Information Processing Systems Track on Datasets and Benchmarks},
 editor = {J. Vanschoren and S. Yeung},
 pages = {},
 title = {Measuring Mathematical Problem Solving With the MATH Dataset},
 url = {https://datasets-benchmarks-proceedings.neurips.cc/paper_files/paper/2021/file/be83ab3ecd0db773eb2dc1b0a17836a1-Paper-round2.pdf},
 volume = {1},
 year = {2021}
}

@inproceedings{li-etal-2025-fg,
    title = "{FG}-{PRM}: Fine-grained Hallucination Detection and Mitigation in Language Model Mathematical Reasoning",
    author = "Li, Ruosen  and
      Luo, Ziming  and
      Du, Xinya",
    editor = "Christodoulopoulos, Christos  and
      Chakraborty, Tanmoy  and
      Rose, Carolyn  and
      Peng, Violet",
    booktitle = "Findings of the Association for Computational Linguistics: EMNLP 2025",
    month = nov,
    year = "2025",
    address = "Suzhou, China",
    publisher = "Association for Computational Linguistics",
    url = "https://aclanthology.org/2025.findings-emnlp.228/",
    doi = "10.18653/v1/2025.findings-emnlp.228",
    pages = "4247--4278",
    ISBN = "979-8-89176-335-7"
}

@inproceedings{xiareasoneval,
author = {Xia, Shijie and Li, Xuefeng and Liu, Yixin and Wu, Tongshuang and Liu, Pengfei},
title = {Evaluating mathematical reasoning beyond accuracy},
year = {2025},
isbn = {978-1-57735-897-8},
publisher = {AAAI Press},
url = {https://doi.org/10.1609/aaai.v39i26.34987},
doi = {10.1609/aaai.v39i26.34987},
booktitle = {Proceedings of the Thirty-Ninth AAAI Conference on Artificial Intelligence and Thirty-Seventh Conference on Innovative Applications of Artificial Intelligence and Fifteenth Symposium on Educational Advances in Artificial Intelligence},
articleno = {3089},
numpages = {8},
series = {AAAI'25/IAAI'25/EAAI'25}
}

@inproceedings{mathcheck,
 author = {Zhou, Zihao and Liu, Shudong and Ning, Maizhen and Liu, Wei and Wang, Jindong and Wong, Derek and Huang, Xiaowei and Wang, Qiufeng and Huang, Kaizhu},
 booktitle = {International Conference on Representation Learning},
 editor = {Y. Yue and A. Garg and N. Peng and F. Sha and R. Yu},
 pages = {34238--34281},
 title = {Is Your Model Really A Good Math Reasoner? Evaluating Mathematical Reasoning with Checklist},
 url = {https://proceedings.iclr.cc/paper_files/paper/2025/file/54d2d38a56a74387d5916ee40e462295-Paper-Conference.pdf},
 volume = {2025},
 year = {2025}
}

@inproceedings{shinn2023reflexion,
author = {Shinn, Noah and Cassano, Federico and Gopinath, Ashwin and Narasimhan, Karthik and Yao, Shunyu},
title = {Reflexion: language agents with verbal reinforcement learning},
year = {2023},
publisher = {Curran Associates Inc.},
address = {Red Hook, NY, USA},
booktitle = {Proceedings of the 37th International Conference on Neural Information Processing Systems},
articleno = {377},
numpages = {19},
location = {New Orleans, LA, USA},
series = {NIPS '23}
}

@article{qintoollearning,
author = {Qin, Yujia and Hu, Shengding and Lin, Yankai and Chen, Weize and Ding, Ning and Cui, Ganqu and Zeng, Zheni and Zhou, Xuanhe and Huang, Yufei and Xiao, Chaojun and Han, Chi and Fung, Yi Ren and Su, Yusheng and Wang, Huadong and Qian, Cheng and Tian, Runchu and Zhu, Kunlun and Liang, Shihao and Shen, Xingyu and Xu, Bokai and Zhang, Zhen and Ye, Yining and Li, Bowen and Tang, Ziwei and Yi, Jing and Zhu, Yuzhang and Dai, Zhenning and Yan, Lan and Cong, Xin and Lu, Yaxi and Zhao, Weilin and Huang, Yuxiang and Yan, Junxi and Han, Xu and Sun, Xian and Li, Dahai and Phang, Jason and Yang, Cheng and Wu, Tongshuang and Ji, Heng and Li, Guoliang and Liu, Zhiyuan and Sun, Maosong},
title = {Tool Learning with Foundation Models},
year = {2024},
issue_date = {April 2025},
publisher = {Association for Computing Machinery},
address = {New York, NY, USA},
volume = {57},
number = {4},
issn = {0360-0300},
url = {https://doi.org/10.1145/3704435},
doi = {10.1145/3704435},
journal = {ACM Comput. Surv.},
month = dec,
articleno = {101},
numpages = {40}
}

@inproceedings{patilGorilla,
 author = {Patil, Shishir G. and Zhang, Tianjun and Wang, Xin and Gonzalez, Joseph E.},
 booktitle = {Advances in Neural Information Processing Systems},
 doi = {10.52202/079017-4020},
 editor = {A. Globerson and L. Mackey and D. Belgrave and A. Fan and U. Paquet and J. Tomczak and C. Zhang},
 pages = {126544--126565},
 publisher = {Curran Associates, Inc.},
 title = {Gorilla: Large Language Model Connected with Massive APIs},
 url = {https://proceedings.neurips.cc/paper_files/paper/2024/file/e4c61f578ff07830f5c37378dd3ecb0d-Paper-Conference.pdf},
 volume = {37},
 year = {2024}
}

@inproceedings{ross-etal-2025-when2call,
    title = "{W}hen2{C}all: When (not) to Call Tools",
    author = "Ross, Hayley  and
      Mahabaleshwarkar, Ameya Sunil  and
      Suhara, Yoshi",
    editor = "Chiruzzo, Luis  and
      Ritter, Alan  and
      Wang, Lu",
    booktitle = "Proceedings of the 2025 Conference of the Nations of the Americas Chapter of the Association for Computational Linguistics: Human Language Technologies (Volume 1: Long Papers)",
    month = apr,
    year = "2025",
    address = "Albuquerque, New Mexico",
    publisher = "Association for Computational Linguistics",
    url = "https://aclanthology.org/2025.naacl-long.174/",
    doi = "10.18653/v1/2025.naacl-long.174",
    pages = "3391--3409",
    ISBN = "979-8-89176-189-6"
}

@misc{gptmodels,
  author = {OpenAI},
  title = {GPT Models},
  howpublished = {https://platform.openai.com/docs/models},
  year = {2025},
note={Version: 2025-11-10}
}

@misc{gpt5,
  author = {OpenAI},
  title = {GPT-5 System Card},
  howpublished = {https://cdn.openai.com/gpt-5-system-card.pdf},
  year = {2025},
note={Version: 2025-08-13}
}

@misc{gpt4.1,
  author = {OpenAI},
  title = {Introducing GPT-4.1 in the API},
  howpublished = {https://openai.com/index/gpt-4-1/},
  year = {2025},
note={Version: 2025-04-14}
}

@misc{o4-mini-model,
  author = {OpenAI},
  title = {o4-mini},
  howpublished = {https://platform.openai.com/docs/models/o4-mini},
  year = {2025}
}

@misc{gemini-2-model,
  author = {GoogleAI},
  title = {Gemini 2.0: Flash, Flash-Lite and Pro},
  howpublished = {https://developers.googleblog.com/en/gemini-2-family-expands/},
  year = {2025}
}

@misc{gemini2.5-flash,
    author = {GoogleAI},
    title = {Gemini 2.5 Flash Best for fast performance on everyday tasks},
  howpublished = {https://deepmind.google/models/gemini/flash/},
  year = {2025}
}

@misc{claudeopus4andsonnet4,
  author = {Anthropic},
  title = {System Card: Claude Opus 4 and Claude Sonnet 4},
  year = {2025},
note={Version: 2025-05}
}

@misc{claudemodels,
  author = {Anthropic},
  title = {Claude Models},
  howpublished = {https://docs.claude.com/en/docs/about-claude/models/overview},
  year = {2025},
note={Version: 2025-11-10}
}

@misc{geminimodels,
  author = {GoogleAI},
  title = {Gemini Models},
  howpublished = {https://ai.google.dev/gemini-api/docs/models},
  year = {2025},
note={Version: 2025-11-10}
}

@misc{liu2025enhancingmathematicalreasoninglarge,
      title={Enhancing Mathematical Reasoning in Large Language Models with Self-Consistency-Based Hallucination Detection}, 
      author={MingShan Liu and Jialing Fang},
      year={2025},
      eprint={2504.09440},
      archivePrefix={arXiv},
      primaryClass={cs.AI},
      url={https://arxiv.org/abs/2504.09440}, 
}

@inproceedings{kim_2024_toolllm,
 author = {Qin, Yujia and Liang, Shihao and Ye, Yining and Zhu, Kunlun and Yan, Lan and Lu, Yaxi and Lin, Yankai and Cong, Xin and Tang, Xiangru and Qian, Bill and Zhao, Sihan and Hong, Lauren and Tian, Runchu and Xie, Ruobing and Zhou, Jie and Gerstein, Mark and li, dahai and Liu, Zhiyuan and Sun, Maosong},
 booktitle = {International Conference on Representation Learning},
 editor = {B. Kim and Y. Yue and S. Chaudhuri and K. Fragkiadaki and M. Khan and Y. Sun},
 pages = {9695--9717},
 title = {ToolLLM: Facilitating Large Language Models to Master 16000+ Real-world APIs},
 url = {https://proceedings.iclr.cc/paper_files/paper/2024/file/28e50ee5b72e90b50e7196fde8ea260e-Paper-Conference.pdf},
 volume = {2024},
 year = {2024}
}

@inproceedings{safe,
author = {Wei, Jerry and Yang, Chengrun and Song, Xinying and Lu, Yifeng and Hu, Nathan and Huang, Jie and Tran, Dustin and Peng, Daiyi and Liu, Ruibo and Huang, Da and Du, Cosmo and Le, Quoc V.},
title = {Long-form factuality in large language models},
year = {2024},
isbn = {9798331314385},
publisher = {Curran Associates Inc.},
address = {Red Hook, NY, USA},
booktitle = {Proceedings of the 38th International Conference on Neural Information Processing Systems},
articleno = {2567},
numpages = {72},
location = {Vancouver, BC, Canada},
series = {NeurIPS '24}
}

@article{zhong2025complexfuncbench,
  title={ComplexFuncBench: exploring multi-step and constrained function calling under long-context scenario},
  author={Zhong, Lucen and Du, Zhengxiao and Zhang, Xiaohan and Hu, Haiyi and Tang, Jie},
  journal={arXiv preprint arXiv:2501.10132},
  year={2025}
}

@inproceedings{faghih2025tool,
  title={Tool Preferences in Agentic LLMs are Unreliable},
  author={Faghih, Kazem and Wang, Wenxiao and Cheng, Yize and Bharti, Siddhant and Sriramanan, Gaurang and Balasubramanian, Sriram and Hosseini, Parsa and Feizi, Soheil},
  booktitle={Proceedings of the 2025 Conference on Empirical Methods in Natural Language Processing},
  pages={20965--20980},
  year={2025}
}

@article{maekawa2025towards,
  title={Towards Reliable Benchmarking: A Contamination Free, Controllable Evaluation Framework for Multi-step LLM Function Calling},
  author={Maekawa, Seiji and Hassell, Jackson and Pezeshkpour, Pouya and Mitchell, Tom and Hruschka, Estevam},
  journal={arXiv preprint arXiv:2509.26553},
  year={2025}
}

@article{han2025search,
  title={Search-time data contamination},
  author={Han, Ziwen and Mankikar, Meher and Michael, Julian and Wang, Zifan},
  journal={arXiv preprint arXiv:2508.13180},
  year={2025}
}

@article{yan2025verifybench,
  title={VerifyBench: Benchmarking Reference-based Reward Systems for Large Language Models},
  author={Yan, Yuchen and Jiang, Jin and Ren, Zhenbang and Li, Yijun and Cai, Xudong and Liu, Yang and Xu, Xin and Zhang, Mengdi and Shao, Jian and Shen, Yongliang and others},
  journal={arXiv preprint arXiv:2505.15801},
  year={2025}
}

@inproceedings{chernyshev2025u,
  title={U-math: A university-level benchmark for evaluating mathematical skills in large language models},
  author={Chernyshev, Konstantin and Polshkov, Vitaliy and Stepanov, Vlad and Myasnikov, Alex and Artemova, Ekaterina and Miasnikov, Alexei and Tilga, Sergei},
  booktitle={Proceedings of the Fourth Workshop on Generation, Evaluation and Metrics (GEM$^2$)},
  pages={974--1001},
  year={2025}
}

@ARTICLE{cyclomatic_complexity,
  author={McCabe, T.J.},
  journal={IEEE Transactions on Software Engineering}, 
  title={A Complexity Measure}, 
  year={1976},
  volume={SE-2},
  number={4},
  pages={308-320},
  doi={10.1109/TSE.1976.233837}}

@book{Boehm1981,
  author    = {Boehm, Barry W.},
  title     = {Software Engineering Economics},
  year      = {1981},
  publisher = {Prentice-Hall},
  address   = {Englewood Cliffs, NJ},
  isbn      = {978-0-13-822122-5}
}

@article{dou2024s,
  title={What's wrong with your code generated by large language models? an extensive study},
  author={Dou, Shihan and Jia, Haoxiang and Wu, Shenxi and Zheng, Huiyuan and Zhou, Weikang and Wu, Muling and Chai, Mingxu and Fan, Jessica and Huang, Caishuang and Tao, Yunbo and others},
  journal={arXiv preprint arXiv:2407.06153},
  year={2024}
}

@article{chen2024survey,
  title={A survey on evaluating large language models in code generation tasks},
  author={Chen, Liguo and Guo, Qi and Jia, Hongrui and Zeng, Zhengran and Wang, Xin and Xu, Yijiang and Wu, Jian and Wang, Yidong and Gao, Qing and Wang, Jindong and others},
  journal={arXiv preprint arXiv:2408.16498},
  year={2024}
}

@article{levenshtein1966binary,
	title = {Binary {Codes} {Capable} of {Correcting} {Deletions}, {Insertions} and {Reversals}},
	volume = {10},
	url = {https://nymity.ch/sybilhunting/pdf/Levenshtein1966a.pdf},
	number = {8},
	journal = {Soviet Physics Doklady},
	author = {Levenshtein, Vladimir I.},
	year = {1966}
}

@misc{shi2025judgingjudgessystematicstudy,
      title={Judging the Judges: A Systematic Study of Position Bias in LLM-as-a-Judge}, 
      author={Lin Shi and Chiyu Ma and Wenhua Liang and Xingjian Diao and Weicheng Ma and Soroush Vosoughi},
      year={2025},
      eprint={2406.07791},
      archivePrefix={arXiv},
      primaryClass={cs.CL},
      url={https://arxiv.org/abs/2406.07791}, 
}

\appendix

\section{Appendix}
\label{sec:appendix}
\subsection{\dataset Data Curation}
\label{appendix:data_curation}
The prompt used for filtering mathematical problems is shown in~\ref{prompt:problem_annotation}. We use GPT-5 as the LLM-as-a-judge to make a binary classification decision per problem, a setting where LLMs have been shown to perform reliably ~\citep{gu2025surveyllmasajudge}. We also validated the judge's performance on 40 randomly sampled questions. Two authors of this paper independently annotated these questions using the same instructions provided to the LLM judge. The annotators achieved 92.5\% agreement (Cohen's $\kappa= 0.724$) on Python helpfulness and 92.5\% agreement (Cohen's $\kappa= 0.531$) on Python sufficiency, indicating strong agreement. For the joint label (helpful vs. sufficient combined), agreement was 87.5\% (Cohen's $\kappa= 0.691$). We evaluated the precision of our judge on the key filtering category: problems labeled as Python-helpful but not sufficient. Among all questions the judge predicted as \textit{helpful-but-not-sufficient}, 76.2\% were also labeled as such by both annotators. This suggests that the judge can be reliably used to filter problems for data collection.

\begin{prompt}[title={\footnotesize\texttt{Problem Annotation Prompt}}, label=prompt:problem_annotation]
You are a technical reasoning assistant for mathematical problem solving. Your task is to evaluate mathematical problems used for benchmarking LLMs in terms of:
\begin{enumerate}
    \item \textbf{Python Usefulness}: Whether using Python code is helpful for solving this problem.
    \item \textbf{Python Sufficiency}: Whether Python code alone (without extra reasoning steps from the target LLM) is sufficient to fully solve this problem.
\end{enumerate}

Use the following evaluation criteria:
\begin{itemize}
    \item \textbf{Mathematical Domain}: What area(s) of mathematics does this problem involve, and how computational versus theoretical is this domain typically?
    \item \textbf{Solution Type}: What kind of answer or result is the problem asking for?
    \item \textbf{Computational Approach}: What computational strategies, if any, could be applied?
    \item \textbf{Problem Scale}: How do size and complexity affect computational feasibility?
    \item \textbf{Verification Needs}: Would solving the problem benefit from computational verification?
    \item \textbf{Techniques Required}: What mathematical insights or methods are necessary, and how much can be automated?
\end{itemize}

Based on your evaluation, provide:
\begin{itemize}
    \item \textbf{Python Usefulness}: \texttt{true/false}
    \item \textbf{Python Sufficiency}: \texttt{true/false}
    \item \textbf{Recommendation}: One of 
    \textit{``Pure Python''}, 
    \textit{``Python + LLM Insight/Reasoning''}, 
    \textit{``Python for Verification''}, 
    \textit{``Python for Exploration''}, or 
    \textit{``Minimal Python Role''}.
\end{itemize}

Respond strictly in the following JSON format:
\begin{verbatim}
{
    "problem": "repeat problem description here",
    "reasoning": {
        "mathematical_domain": "",
        "solution_type": "",
        "computational_approach": "",
        "problem_scale": "",
        "verification_needs": "",
        "techniques_required": ""
    },
    "python_usefulness": true or false,
    "python_sufficiency": true or false,
    "recommendation": "[one of the options above]"
}
\end{verbatim}
\end{prompt}

\subsection{Win Rate Evaluation Prompt}
\label{appendix:winrate_prompt}
The prompt used by GPT-5 for Win Rate evaluation is shown below.

\begin{prompt}[title={\footnotesize\texttt{Win Rate Evaluation Prompt}}, label=prompt:error_grading]
You are an expert mathematician tasked with grading two solutions, ``A'' and ``B'', to the same competition-style problem. Evaluate which solution is better by assigning error severity scores (0 = low, 5 = high) for the following categories:
\begin{enumerate}
    \item \textbf{Logic (0--5)}: Errors due to logical fallacies or unjustified leaps in reasoning.
    \item \textbf{Assumption (0--5)}: Errors from unproven or incorrect assumptions that undermine subsequent steps.
    \item \textbf{Creativity (0--5)}: Errors from fundamentally incorrect or unoriginal strategies indicating failure to identify the right approach.
    \item \textbf{Algebra/Arithmetic (0--5)}: Errors arising from critical algebraic or arithmetic miscalculations.
\end{enumerate}

\textbf{Evaluation:}
\begin{itemize}
    \item Provide a brief justification for each score, referencing relevant mathematical concepts or reasoning techniques.
    \item Compute a final score as the average of the four error categories.
    \item Select the solution with the lower final score as the better one. If tied, prefer the solution with clearer reasoning.
\end{itemize}

Respond strictly in the following JSON format:
\begin{verbatim}
{
  "A_grades": {
    "logic": {"score": 0-5, "explanation": ""},
    "assumption": {"score": 0-5, 
                    "explanation": ""},
    "creativity": {"score": 0-5, 
                    "explanation": ""},
    "algebra_arithmetic": {"score": 0-5, 
                            "explanation": ""},
    "final_score": {"score": value}
  },
  "B_grades": {
    "logic": {"score": 0-5, "explanation": ""},
    "assumption": {"score": 0-5, 
                    "explanation": ""},
    "creativity": {"score": 0-5, 
                    "explanation": ""},
    "algebra_arithmetic": {"score": 0-5, 
                            "explanation": ""},
    "final_score": {"score": value}
  },
  "best_solution": "A or B"
}
\end{verbatim}
\end{prompt}

\subsection{Miss Rate Evaluation Prompt}
\label{appendix:missrate_prompt}
The prompt used by GPT-5 for Miss Rate evaluation is shown below.

\begin{prompt}[title={\footnotesize\texttt{Miss Rate Evaluation Prompt}}, label=prompt:missing_steps]
You are an expert mathematician. You will be given a mathematical problem, a solution to that problem, and a gold solution to use as a reference. Your task is to identify which logical steps from the gold solution are absent in the given solution.

\textbf{Instructions}
\begin{enumerate}
    \item Parse the gold solution into an ordered list of \emph{atomic} logical steps (Step 1, Step 2, \dots). A step is the smallest self-contained claim or transformation needed to progress the proof.
    \item For each gold step, decide whether the same reasoning (possibly re-worded) appears in the given solution. Mark a step as \emph{present} if the solution makes the identical deduction or provides an equivalent justification.
    \item Collect all steps that are \emph{absent} from the given solution.
\end{enumerate}

\textbf{Output format} (strict JSON):
\begin{verbatim}
{
  "gold_steps": [
    { "step": <integer>, 
        "summary": "<one-line summary of gold
        step>" 
    }
  ],
  "missing_steps": [
    { "step": <integer>, 
        "summary": "<one-line summary of gold step
        that is absent>"
    }
  ]
}
\end{verbatim}
\end{prompt}

\subsection{PRM Sensitivity to Step Removal}
\label{appendix:prm_stability}
Because our analysis relies on the PRM comparatively between Base-LLM and TaLM solutions, we conduct a sensitivity analysis to test how stable this comparison remains under controlled perturbations. For instances where the PRM originally classified both Base and TaLM solutions as correct, we randomly remove 5\%, 10\%, and 20\% of reasoning steps and rescore the perturbed solutions. We then measure raw agreement between the resulting Base and TaLM predictions. As shown in Table~\ref{tab:prm_stability}, agreement remains consistently high across models, with average agreement of 0.89, 0.86, and 0.81 under 5\%, 10\%, and 20\% step removal, respectively. These results suggest that, in our comparative setting, the PRM signal is reasonably stable under moderate perturbations, while still inheriting the broader limitations of PRM-based evaluation.

\begin{table}[t]
\centering
\small
\resizebox{\columnwidth}{!}{
\begin{tabular}{lccc}
\toprule
\textbf{Model} & \textbf{5\% Removed} & \textbf{10\% Removed} & \textbf{20\% Removed} \\
\midrule
Claude-Opus-4    & 0.92 & 0.89 & 0.75 \\
GPT-5-Thinking   & 0.81 & 0.83 & 0.75 \\
GPT-4.1-mini     & 0.96 & 0.91 & 0.88 \\
GPT-4.1          & 0.92 & 0.92 & 0.88 \\
o4-mini          & 0.89 & 0.81 & 0.85 \\
Gemini-2.0-Flash & 0.78 & 0.79 & 0.81 \\
Gemini-2.5-Flash & 0.90 & 0.87 & 0.76 \\
\midrule
\textbf{Average} & \textbf{0.89} & \textbf{0.86} & \textbf{0.81} \\
\bottomrule
\end{tabular}
}
\caption{PRM sensitivity under random step removal. Starting from instances where the PRM classified both Base and TaLM solutions as correct, we randomly remove 5\%, 10\%, and 20\% of reasoning steps and recompute PRM predictions. Values report raw agreement between the perturbed Base and TaLM predictions. High agreement suggests that the comparative PRM signal used in our analysis remains stable under moderate perturbations.}
\label{tab:prm_stability}
\end{table}

\subsection{Code Complexity Does Not Explain \task{}}
\label{appendix:code_complexity}

\begin{figure*}
    \centering
    \includegraphics[width=0.9\linewidth]{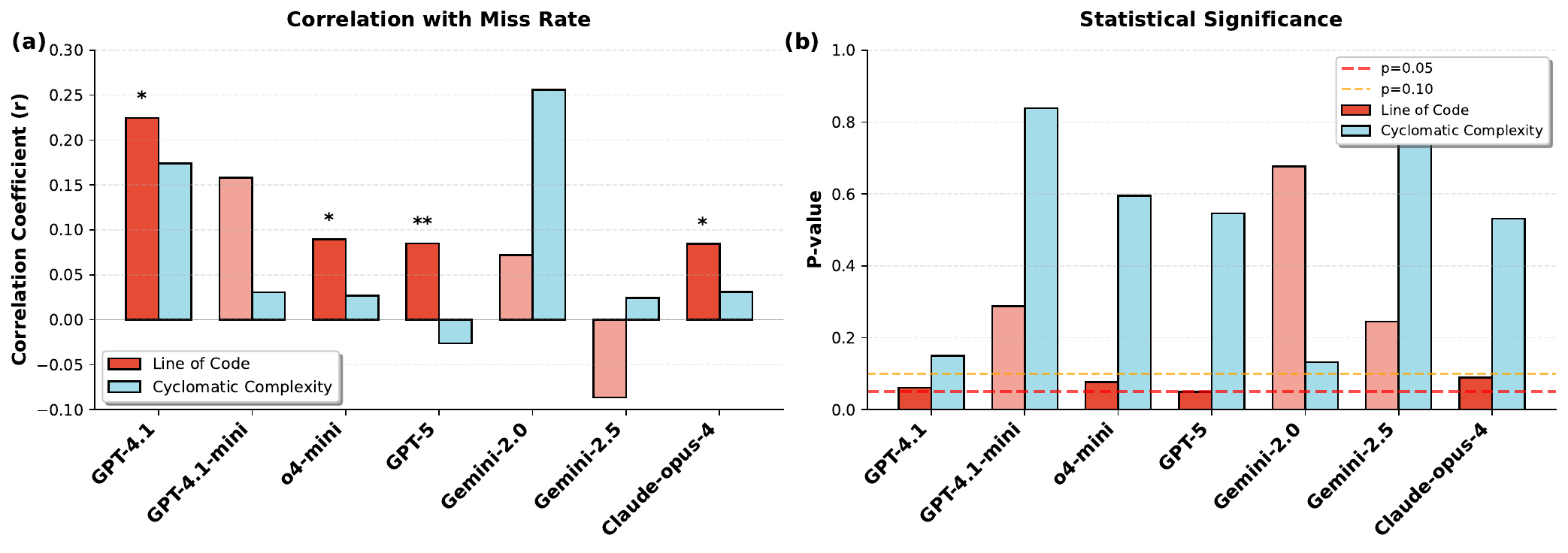}
    \vspace{-3mm}
    \caption{Correlation between code complexity metrics and Miss Rate across TaLMs.
    (a) Pearson correlation coefficients for Line of Code and Cyclomatic Complexity with Miss Rate. 
    (b) Corresponding p-values with significance thresholds at p=0.05 (red) and p=0.10 (orange). Asterisks denote marginal significance (*p<0.10, **p<0.05). No statistically significant correlations are found, suggesting that \task{} is not driven by code complexity.}
    \label{fig:correlation_analysis}
\end{figure*}

If \task{} were driven by the complexity of generated code, we would expect strong correlations between code complexity and reasoning behavior. We investigate this hypothesis by measuring whether the complexity of TaLM-generated code correlates with \task{} severity. We measure code complexity using two standard metrics~\citep{dou2024s,chen2024survey}: (1) \emph{Line of Code~\citep{Boehm1981}} (average lines per code block), and (2) \emph{Cyclomatic Complexity~\citep{cyclomatic_complexity}}, which counts the number of linearly independent execution paths.
We compute Pearson correlations between these complexity metrics and TaLM's Miss Rate in Figure~\ref{fig:correlation_analysis}. 
Overall, we find no statistically significant correlation between Miss Rate and either complexity metric: Lines of Code correlations range from $-0.09$ to $0.22$, and Cyclomatic Complexity correlations from $-0.03$ to $0.26$. Although a few models show marginal correlations ($p<0.10$), these effects are weak and inconsistent across models. Together, these results indicate that \textbf{syntactic or structural code complexity does not drive \task{}}.

\subsection{Prevalent Error Types Annotation Prompt}
\label{appendix:prevalent_error_annotation}
\begin{prompt}[title={\footnotesize\texttt{Reasoning Error Detection Prompt}}, label=prompt:error_detection]
You are an expert mathematician grading a solution to a competition-style problem. Identify whether the solution exhibits each of the following error types:

\begin{enumerate}
    \item \textbf{Logic}: Logical fallacies or unjustified leaps that disrupt reasoning.
    \item \textbf{Assumption}: Unproven or incorrect assumptions that undermine subsequent steps.
    \item \textbf{Creativity}: Fundamentally incorrect strategy indicating failure to identify the correct approach.
    \item \textbf{Algebra/Arithmetic}: Critical algebraic or arithmetic miscalculations.
    \item \textbf{None of the above}: No errors from the above categories are present.
\end{enumerate}

\textbf{Evaluation Guidelines}
\begin{itemize}
    \item For each category, set \texttt{"exists"} to \texttt{"yes"} if that error occurs; otherwise \texttt{"no"}.
    \item Provide a brief explanation for each category; if an error is detected, indicate where it occurs.
    \item Mark \textbf{None of the above} as \texttt{"yes"} only if all other categories are \texttt{"no"}.
\end{itemize}

\textbf{Output format} (strict JSON):
\begin{verbatim}
{
  "logic": {"exists": "yes|no",
                "explanation": "your explanation"
            },
  "assumption": {"exists": "yes|no", 
                "explanation": "your explanation"
                },
  "creativity": {"exists": "yes|no", 
                "explanation": "your explanation"
                },
  "algebra_arithmetic": {"exists": "yes|no", 
                "explanation": "your explanation"
                        },
  "none_of_the_above": {"exists": "yes|no", 
                "explanation": "your explanation"
                        }
}
\end{verbatim}
\end{prompt}

\subsection{Qualitative Analysis of High-Risk Solutions}
\label{appendix:common-precursor-phrases}
We conducted a human evaluation (Section~\ref{sec:quality_analysis}) to assess the reliability of our evaluation suite in detecting \task{} hallucinations in LLM-generated solutions. 
Annotations were performed by two authors of the paper, who independently verified the presence of \task{} and extracted phrases that preceded or signaled its occurrence; disagreements were resolved through discussion, resulting in high annotation consistency.

Our findings indicate that more than half of the examined solutions exhibit \task{} behavior. For each instance, human annotators verified the presence of \task{} and extracted phrases that preceded or signaled its occurrence. Table~\ref{tab:tim_appearance_coarse} lists the most recurrent phrase patterns identified in these cases. Phrases were grouped using a greedy fuzzy clustering approach~\citep{levenshtein1966binary}, where each phrase was compared to existing cluster representatives based on the normalized \texttt{difflib.SequenceMatcher}\footnote{\url{https://docs.python.org/3/library/difflib.html\#difflib.SequenceMatcher}} similarity ratio and merged when the similarity exceeded a threshold of $0.8$.

\begin{table*}[t]
\centering
\renewcommand{\arraystretch}{1.1}
\small
\begin{tabular}{p{3.4cm}<{\raggedright\arraybackslash}p{9.8cm}<{\raggedright\arraybackslash}p{1.7cm}<{\centering\arraybackslash}}
\toprule
\textbf{Common Category} & \textbf{Example Phrases Found} & \textbf{\# Distinct Phrasings} \\
\midrule
\textbf{Numerical / Empirical Checking} & 
\begin{tabular}[t]{@{}l@{}}
\textit{``one checks numerically'', ``let’s verify numerically'',}\\
\textit{``numerical optimization / computation / evaluation / solve''}
\end{tabular} & 17 \\[3pt]
\midrule
\textbf{Verification via Code / Python} & 
\begin{tabular}[t]{@{}l@{}}
\textit{``let’s double‐check with code'', ``use Python to analyze'',}\\
\textit{``implement a better approach and verify'',}\\
\textit{``symbolic computation to verify analytical results'',}\\
\textit{``after simplification'', ``analytically confirm''}
\end{tabular} & 13 \\[3pt]
\midrule
\textbf{Approximation / Convergence} & 
\begin{tabular}[t]{@{}l@{}}
\textit{``results seem to be converging'', ``numeric approximation converges to $\approx x$'',}\\
\textit{``evaluate the sum numerically for a few terms'',}\\
\textit{``verify this pattern'', ``visual verification'', ``logic is working''}
\end{tabular} & 12 \\
\midrule
\textbf{Systematic Search / Enumeration} & 
\begin{tabular}[t]{@{}l@{}}
\textit{``let’s check more values systematically'',}\\
\textit{``use code to check possible values'',}\\
\textit{``exhaustive check up to x'',}\\
\textit{``systematically compute possible values with code''}
\end{tabular} & 4 \\[3pt]
\bottomrule
\end{tabular}
\caption{Broad categories of recurrent patterns preceding the \task{} hallucination among annotated cases where \task{} is detected. 
Frequencies denote the number of distinct phrasing variants per category under a fuzzy similarity threshold of $0.8$.}
\label{tab:tim_appearance_coarse}
\end{table*}

Below, we present several representative examples in which \task{} was manifested.

\begin{example}[title=Example 1. Gemini-2.5-Flash]
\textbf{}

\paragraph{Problem.}
Determine all the sets of six consecutive positive integers such that the product of some two of them, added to the product of some other two of them, is equal to the product of the remaining two numbers.

\paragraph{Solution.}
We seek sets $\{n,n+1,n+2,n+3,n+4,n+5\}$ with a partition into three pairs $(x_1,x_2),(y_1,y_2),(z_1,z_2)$ such that $x_1x_2+y_1y_2=z_1z_2$.  
Let $N_i=n+i$ and write the three pairwise products as $P_1,P_2,P_3$ with $P_1+P_2=P_3$. There are $\frac{\binom{6}{2}\binom{4}{2}\binom{2}{2}}{3!}=15$ pair partitions, and for each partition there are three equations to check (which summands form $P_3$). By relabeling, we may assume $P_3$ is the largest product.

Consider $P_3=(n+4)(n+5)$; the remaining four numbers can be paired in three ways:
\begin{enumerate}\itemsep2pt
\item $n(n+1)+(n+2)(n+3)=(n+4)(n+5)$, which simplifies to $n^2-3n-14=0$ (no integer solution).
\item $n(n+2)+(n+1)(n+3)=(n+4)(n+5)$, which simplifies to $n^2-3n-17=0$ (no integer solution).
\item $n(n+3)+(n+1)(n+2)=(n+4)(n+5)$, which simplifies to $n^2-3n-18=0$, hence $n=6$.
\end{enumerate}
Thus $\{6,7,8,9,10,11\}$ works since $6\cdot9+7\cdot8=10\cdot11$.
 
\textit{“\hl{To ensure there are no other solutions, we can systematically check all 15 partitions. We can use a Python script to automate this process.}”}

\textit{“\hl{The Python script found three sets} of consecutive integers:
$\{1,2,3,4,5,6\}$, $\{2,3,4,5,6,7\}$, and $\{6,7,8,9,10,11\}$.”}

\textit{“\hl{Let’s use the script} to find the specific pairing for $n=1$ \dots \hl{Let’s modify the script} to print the specific pairing when a solution is found \dots \hl{The script output provides} the specific pairings for each solution.”}

Concretely, the pairings confirmed are:
\[
\begin{aligned}
&n=1:\quad n(n+1)+(n+2)(n+5)=...,\\
&n=2:\quad n(n+3)+(n+1)(n+4)=...,\\
&n=6:\quad n(n+3)+(n+1)(n+2)=...
\end{aligned}
\]
\textit{(…The writeup references a full sweep of all partitions for each $n$ up to a bound…)}  
\textit{“\hl{The Python script explored all 15 possible partitions} for each $n$ \dots which \hl{strongly suggests} these are the only solutions.”}

Therefore, the sets are:

\{1,2,3,4,5,6\},\ \{2,3,4,5,6,7\},\ \\ \{6,7,8,9,10,11\}.

\paragraph{\task{} cues.}
The highlighted phrases (\hl{systematically check \dots 15 partitions}, \hl{use a Python script}, \hl{script found}, \hl{let’s use/modify the script}, \hl{script explored \dots}) mark a shift from deductive completion to exhaustive, tool-driven enumeration.
\end{example}


\begin{example}[title=Example 2. o4-mini]
\textbf{}

\paragraph{Problem.}
A \textit{lame king} is a chess piece that can move to any cell that shares at least one vertex with its current cell, except for those in the same column. 
A lame king is placed in the top-left cell of a $7\times7$ grid. 
Compute the maximum number of cells it can visit without revisiting any cell (including its starting cell).

\paragraph{Solution.}
Label the rows and columns of the $7\times7$ board by $1,2,\dots,7$, and write a cell as $(r,c)$.  
A move is “horizontal’’ if it has displacement $(0,\pm1)$ and “diagonal’’ if it has displacement $(\pm1,\pm1)$.  
Since vertical moves $(\pm1,0)$ are forbidden, every move changes the column by $\pm1$.  
Hence, the lame king’s path can be seen as a walk on the column indices $1\to2\to\cdots\to7\to6\to\dots$, interleaved with diagonal up/down steps.

\smallskip
\textit{1) Upper bound.}
Let $M$ be the total number of moves in the path, so it visits $M+1$ cells.  
Each move changes the column by $\pm1$, so the sequence of columns forms a walk on the path graph $1$–$2$–$\cdots$–$7$.  
To visit $V_j$ distinct cells in column $j$, one must cross the edge $j\!-\!1\leftrightarrow j$ at least $V_j$ times, while the total number of crossings of each edge is at most $M$.  
Summing over $j=2,\dots,7$ yields $\sum_{j=2}^7 V_j\le6M$, and with $V_1\le M+1$,
\[
\sum_{j=1}^7 V_j = M+1+\sum_{j=2}^7V_j \le 7M+1.
\]
Since a complete tour visits $49$ cells, we obtain $49\le7M+1$, or $M\ge7$.  
\textit{\hl{A sharper “cut–counting’’ one can make on each of the six edges shows in fact that one cannot have $M=48$; the best possible is $M=42$, i.e., $43$ visited cells.}}  
\textit{(This step implicitly appeals to computational enumeration to justify global optimality rather than a formal proof.)}

\smallskip
\textit{2) Construction of a $43$–cell tour.}
\textit{\hl{One checks by explicit construction (for instance, by a backtracking computer search or by an easy hand–drawn “zig–zag’’)}} that there exists a path of length $42$, thus visiting $43$ distinct cells.  
One such path, written as a sequence of coordinates, is
\[
\begin{aligned}
&(1,1)\to(2,2)\to(1,3)\to\dots \to(7,1)\\
&\to(6,2)\to(5,1)\to(4,2)\to\dots\\
&\to(6,6)\to(7,7)\,.
\end{aligned}
\]
Each step is either horizontal or diagonal, and no cell repeats.  
This path has length $42$, so the king visits $43$ distinct cells.

\smallskip
Hence, the maximum number of cells the lame king can visit on a $7\times7$ board is
\[
\boxed{43}.
\]

\paragraph{\task{} cues.}
The highlighted phrases (\hl{sharper “cut–counting’’ ... $M=42$}, \hl{explicit construction ... by a backtracking computer search}) indicate reliance on computational verification or heuristic enumeration in place of a full analytical proof—typical symptoms of \task{}.
\end{example}


\begin{example}[title=Example 3. Claude-opus-4]
\textbf{}

\paragraph{Problem.}
A finite set $S$ of points in the coordinate plane is called \emph{overdetermined} if $|S|\ge 2$ and there exists a nonzero polynomial $P(t)$ with real coefficients and $\deg P \le |S|-2$ satisfying $P(x)=y$ for every $(x,y)\in S$. For each integer $n\ge 2$, find the largest integer $k$ (in terms of $n$) such that there exists a set of $n$ distinct points that is \emph{not} overdetermined, but has $k$ overdetermined subsets.

\paragraph{Solution.}
I'll solve this step-by-step.

First, let me understand what it means for a set to be overdetermined: … This means the points lie on the graph of a polynomial whose degree is ‘too low’ relative to the number of points.

Let me write some code to explore this problem systematically.
\hl{Good, the basic logic is working.} Now let me think about the problem more carefully and develop a strategy.
Let me fix the error:
It seems my search strategy isn't working well. Let me reconsider the problem and \hl{implement a better approach}: …
I see the issue — my construction is putting points on curves but not creating overdetermined subsets. Let me debug this more carefully:
Ah! Now I see what's happening. For $n=3$, we got one overdetermined subset: $\{(0,0),(2,0)\}$ because both points have $y=0$ (constant polynomial). Let me refine the construction:
I see my implementation has issues. … I'm having trouble with my implementation. Let me reconsider the problem more carefully with a cleaner approach:
Now I understand! Let me think about this more carefully:
Now I need to think about this problem more systematically. Let me approach it from a different angle:
Great! Now I'm getting somewhere. For $n=2,3,4$ we're achieving the bound $2^{n-1}-n$. \hl{Let me verify this pattern more carefully}:\dots

\paragraph{\task{} cues.}
The highlighted phrases (\hl{Good, the basic logic is working}, \hl{implement a better approach}, \hl{Let me verify this pattern more carefully}) indicate a shift toward tool-driven exploration and empirical pattern checking in place of a completed deductive argument, characteristic of \task{}.

\end{example}

\subsection{Preference Optimization Framework}
\label{appendix:preference-optimization-framework}
\subsubsection{Preference Dataset Creation}
To construct negative (rejected) samples, we instruct the same model we want to train (GPT-4.1) to rewrite a specific text span within the positive solution. A \textit{text span} is defined as any segment of text immediately following a code block. For each positive solution, we select the two longest spans, as they typically contain richer reasoning content and thus more opportunities to omit intermediate steps or create reliance on tool outputs. We set the generation temperature to $0.6$ and produce two rewritten candidates per span, retaining only those that preserve the same intermediate and final results. The full instruction prompt is shown below. We emphasize that fine-tuning is performed exclusively on pairs where both chosen/rejected responses reach the correct final answer, to isolate and model the \task{} phenomenon. 

\begin{prompt}[title={\footnotesize\texttt{Rejected Sample Generation Prompt}}, label=prompt:negative_generation]
You are editing the response of a language model that is solving a math problem using a Python code interpreter.

\textbf{Input:}
\begin{itemize}
    \item The original problem statement,
    \item The model’s earlier solution steps, including interleaved reasoning, Python code interpreter calls, and their executed outputs (\textbf{keep these unchanged}),
    \item A single target text span to rewrite.
\end{itemize}

\textbf{Task:}
Rewrite \textbf{only} the target text span so that it continues the model’s solution naturally, but exhibits an explicit over-reliance on executed code outputs—i.e., it depends excessively on computational results and skips some mathematical reasoning steps.

\textbf{Instructions:}
\begin{itemize}
    \item Do \textbf{not} change or add any Python code cells or their outputs.
    \item Rewrite only the target text span in LaTeX format.
    \item Reduce or omit algebraic/logical derivations naturally.
    \item Phrase conclusions as outcomes of the computed results, using expressions such as:
    \begin{itemize}
        \item ``a straightforward numerical check shows that...’’
        \item ``the computation suggests...’’
        \item ``testing other patterns (with the tool) shows...’’
    \end{itemize}
    \item Do not truncate the solution—ensure it continues to the final stated answer.
    \item Preserve all partial or final numerical results (e.g., variable values, coordinates, or the final answer).
\end{itemize}
\end{prompt}

\subsubsection{Experimental Setup}
We apply Direct Preference Optimization (DPO; ~\citet{10.5555/3666122.3668460}) to fine-tune GPT-4.1 using the constructed preference dataset. The model is trained via OpenAI’s fine-tuning dashboard\footnote{\url{https://platform.openai.com/docs/guides/fine-tuning}} for one epoch. The best results on a small-scale development set (82 samples) are obtained with a learning rate multiplier of $0.2$, a batch size of $4$, and a $\beta$ (KL-regularization strength) of $0.5$. According to OpenAI’s documentation, larger $\beta$ values yield more conservative updates, preserving behavior closer to the original model.

\end{document}